\documentclass[11pt]{article}

\usepackage[final]{acl}

\usepackage{times}
\usepackage{latexsym}

\usepackage[T1]{fontenc}

\usepackage[utf8]{inputenc}

\usepackage{microtype}

\usepackage{inconsolata}

\usepackage{graphicx}
\usepackage{placeins}
\usepackage{float}

\usepackage{amssymb}
\usepackage{booktabs}
\usepackage{multirow}
\usepackage{amsmath}
\usepackage{xcolor}
\usepackage{tabularx}
\usepackage{colortbl}
\usepackage[normalem]{ulem}
\usepackage{pifont}
\newcommand{\xmark}{\ding{55}}
\useunder{\uline}{\ul}{}

\definecolor{gappos}{HTML}{1E7B1E}
\definecolor{gapneg}{HTML}{C0392B}
\newcommand{\accgap}[1]{%
  \dimen0=#1pt\relax
  \ifdim\dimen0>0pt\relax
    {\color{gappos}(+#1)}%
  \else
    {\color{gapneg}(#1)}%
  \fi
}
\newcommand{\accgaprow}[2]{#1\accgap{#2}}
\newcommand{\accpm}[2]{#1$\pm$#2}

\usepackage[normalem]{ulem}
\useunder{\uline}{\ul}{}

\usepackage[most]{tcolorbox}

\title{ObGynLongBench: Revealing the Evidence-to-EHR Gap \\ in Longitudinal EHR Decision-Making}

\author{
Jun Xiang$^{1*}$ \quad
Zhijie Bao$^{1,2*}$ \quad
Rong Hu$^{4\dagger}$ \\
\bfseries Kaizhou Qin$^{3}$ \quad
Wei Chen$^{5\dagger}$ \quad
Zhongyu Wei$^{1,2\dagger}$ \\
$^1$School of Data Science, Fudan University, China \\
$^2$Shanghai Innovation Institute, China \\
$^3$Obstetrics \& Gynecology Hospital of Fudan University, China \\
$^4$Department of Obstetrics, Obstetrics \& Gynecology Hospital of Fudan University, \\
Shanghai Key Lab of Reproduction and Development, \\
Shanghai Key Lab of Female Reproductive Endocrine Related Diseases, Shanghai, China \\
$^5$School of Software Engineering, Huazhong University of Science and Technology, China \\
\texttt{jxiang25@m.fudan.edu.cn, zjbao24@m.fudan.edu.cn, hurong@fudan.edu.cn} \\
\texttt{lemuria\_chen@hust.edu.cn, zywei@fudan.edu.cn}
}

\begin{document}
\maketitle
\begingroup
\renewcommand{\thefootnote}{\fnsymbol{footnote}}
\footnotetext[1]{These authors contributed equally to this work.}
\footnotetext[2]{Corresponding authors.}
\endgroup

\begin{abstract}
The application of large language models (LLMs) to personalized medical assistants has garnered growing interest. However, existing medical benchmarks largely rely on static question answering with pre-selected evidence, leaving unclear whether LLMs can make reliable clinical decisions from real longitudinal electronic health records (EHRs). To bridge this gap, we introduce \textbf{ObGynLongBench}, a rule-grounded long-context EHR benchmark for obstetric and gynecologic decision-making, comprising 1,500 clinical decision-point cases from 976 real pregnancy EHR histories and traceable rules. Each case is anchored to a patient, a pregnancy-timeline point, and a pre-decision information boundary, enabling Evidence-only, Visit-level EHR, and History-level EHR evaluation.
Evaluating 17 LLMs reveals a substantial \textbf{Evidence-to-EHR Gap}: models perform well when evidence is directly provided, but accuracy drops when evidence must be extracted from same-day records or full pre-decision EHR histories. Further analyses identify evidence utilization as a key bottleneck: performance decreases with longer EHR contexts and more complex evidence requirements, and earlier failures often predict later failures within the same patient history. Finally, active-search agents perform best among EHR access strategies, highlighting patient-specific evidence utilization as a central challenge for reliable personalized medical assistants. Resources are available at \url{https://github.com/xiangjun2003/ObgynLongbench}.
\end{abstract}

\section{Introduction}
\label{sec:intro}

Large language models (LLMs) have achieved strong performance on medical examinations and static clinical question answering benchmarks such as MedQA, MedMCQA, PubMedQA, and MultiMedQA~\citep{app11146421,pmlr-v174-pal22a,jin-etal-2019-pubmedqa,Singhal2022LargeLM}. However, real-world personalized medical assistants must go beyond static QA and reason directly over raw patient EHRs. Obstetrics and gynecology provides a representative setting: clinical decisions are often anchored to gestational age and require integrating longitudinal, multi-source evidence accumulated throughout prenatal care~\citep{Goldberg2017CommitteeON,Peahl2020TheEO}.

\begin{figure}[t]
  \centering
  \includegraphics[width=\columnwidth, trim=70pt 60pt 0 0,clip ]{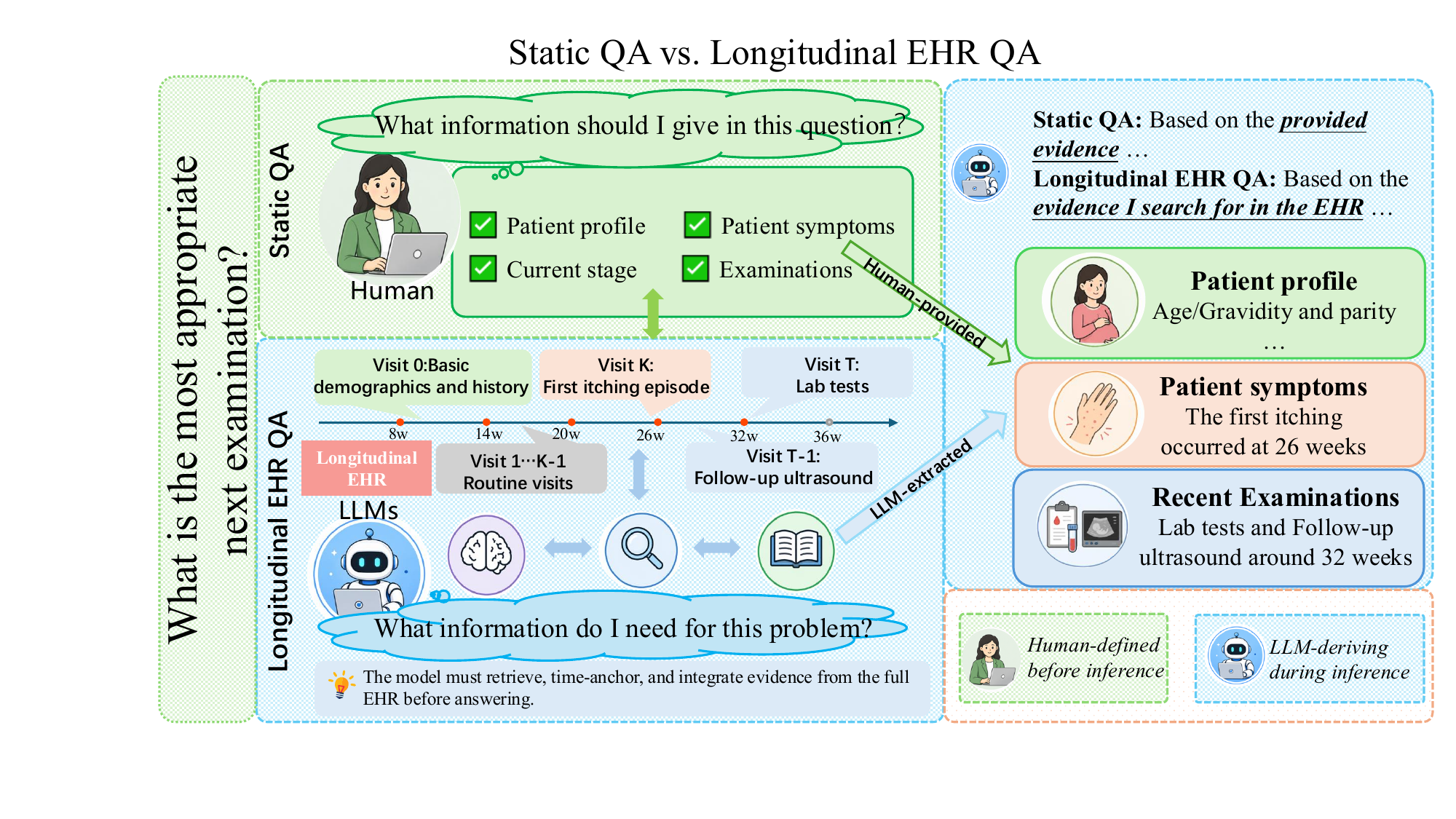}
  \caption{Static QA vs. Longitudinal EHR QA. Static QA uses human-provided decision-relevant evidence, while longitudinal EHR QA requires LLMs to extract patient-specific evidence from the pre-decision EHR history during inference.}
  \label{fig:settings}
\end{figure}


\begin{table*}[t]
\centering
\small
\setlength{\tabcolsep}{4pt}
\renewcommand{\arraystretch}{1.12}
\resizebox{0.9\textwidth}{!}{%
\begin{tabular}{p{5.2cm} p{1.2cm} p{1.8cm} p{2.1cm} p{1.6cm} p{1.4cm}}
\toprule
\textbf{Work} & \textbf{Size} & \textbf{Data Source} & \textbf{Task Format} & \textbf{Input} & \textbf{Guideline} \\
\midrule
\multicolumn{6}{l}{\textit{Obstetrics, gynecology, and maternal-infant QA benchmarks}} \\
\midrule
\begin{tabular}[t]{@{}l@{}}
Exploring ChatGPT in MRCOG \\
\citep{Bachmann2024ExploringTC}
\end{tabular}
& 1,824 
& Exam 
& MCQ 
& Evidence 
& \xmark \\

\begin{tabular}[t]{@{}l@{}}
Comparative Ob/Gyn Board-style QA \\
\citep{ugowski2025ComparativeAO}
\end{tabular}
& 352 
& Exam 
& MCQ 
& Evidence
& \xmark \\

\begin{tabular}[t]{@{}l@{}}
Pregnant Questions\citep{srikanth-etal-2024-pregnant}
\end{tabular}
& 500 
& Patient QA 
& Open 
& Evidence
& \xmark \\

\midrule
\multicolumn{6}{l}{\textit{EHR benchmarks}} \\
\midrule

\begin{tabular}[t]{@{}l@{}}
MedAlign\citep{Fleming2023MedAlignAC}
\end{tabular}
& 983 
& Hospital
& Open 
& History 
& \xmark \\

\begin{tabular}[t]{@{}l@{}}
MIMIC-Instr\citep{NEURIPS2024_62986e0a}
\end{tabular}
& 400K+ 
& Hospital
& Mixed 
& History 
& \xmark \\

\begin{tabular}[t]{@{}l@{}}
TIMER\citep{Cui2025TIMERTI}
\end{tabular}
& 402 / 248 
& Hospital
& Mixed 
& History
& \xmark \\

\midrule

\textbf{ObGynLongBench} (ours)
& 1,500 
& Hospital
& MCQ 
& History
& \checkmark \\

\bottomrule
\end{tabular}
}
\caption{Comparison between existing obstetrics and gynecology QA benchmarks, general EHR benchmarks, and ObGynLongBench. In the ``Input'' column, \textbf{Evidence} denotes the \textbf{Evidence-only} setting, where evidence is provided directly in the question; \textbf{History} denotes the \textbf{History-level EHR} setting, where LLMs reason over longitudinal EHR records. ``Guideline'' indicates grounding in traceable clinical rules from guidelines or textbooks.}
\label{tab:related-benchmarks}
\end{table*}

Existing LLM evaluations in obstetrics and gynecology mainly rely on specialty examinations, board-style questions, or patient-facing question answering~\citep{Bachmann2024ExploringTC,ugowski2025ComparativeAO,srikanth-etal-2024-pregnant}.
As summarized in Table~\ref{tab:related-benchmarks}, these settings largely follow a static QA paradigm analogous to our \textbf{Evidence-only} scope, where relevant clinical facts are provided directly in the question, primarily assessing models' medical knowledge and clinical reasoning ability.
More broadly, general EHR benchmarks have shown that strong performance on static QA does not readily translate to robust performance on EHR-based clinical tasks~\citep{Fleming2023MedAlignAC,Cui2025TIMERTI}.

As illustrated in Figure~\ref{fig:settings}, static QA typically presents evidence in an \textbf{Evidence-only} manner directly in the question, thereby primarily assessing models' medical knowledge.
In real-world obstetric care, however, the relevant evidence is no longer pre-provided in the question, but scattered across the \textbf{History-level EHR}.
Before making a decision, the model must first infer what information is needed for the current clinical problem, retrieve the corresponding evidence from heterogeneous records, time-anchor each piece of evidence along the gestational timeline, and integrate findings across visits, examinations, and clinical notes.

To study this problem, we propose \textbf{ObGynLongBench}, a rule-grounded long-context EHR benchmark for obstetric decision-making along the pregnancy timeline.
The benchmark is built from real longitudinal EHRs of 976 pregnant patients, covering the process from early pregnancy registration to pre-delivery assessment.
Based on textbooks and clinical guidelines, we extract 1,500 traceable clinical decision points across 16 obstetric clinical domains and instantiate them on patient-specific gestational histories.
Each question is anchored to a concrete decision point and associated with a clinical rule and a gold-standard answer.
ObGynLongBench therefore evaluates whether LLMs can locate, time-anchor, and integrate evidence from real longitudinal pregnancy EHRs before making clinical decisions.

ObGynLongBench evaluates LLMs along two controlled axes: evidence complexity and input scope.
For evidence complexity, clinical rules are assigned to three levels: \textbf{L1} single-evidence recognition, \textbf{L2} multi-source integration, and \textbf{L3} long-horizon reasoning over temporally distributed evidence.
The level of a case is determined by the clinical rule used to construct it.
For input scope, each case is tested under three within-case settings: \textbf{Evidence-only} (Evidence), \textbf{Visit-level EHR} (Visit), and \textbf{History-level EHR} (History).
Here, Evidence provides pre-localized key evidence, Visit provides same-day records, and History provides the full pre-decision EHR history, enabling us to measure the \textbf{Evidence-to-EHR Gap} and the sufficiency of visit-level information.

Experimental results reveal a clear \textbf{Evidence-to-EHR Gap}: LLMs perform well under Evidence-only, but accuracy drops substantially when they answer from Visit-level EHR or History-level EHR.

We highlight three main findings.
First, the gap is largely driven by \textbf{evidence utilization}: LLM accuracy decreases as EHR contexts become longer and evidence requirements become more complex.
Second, errors show \textbf{patient-level associations}: LLMs that fail at an earlier decision point are more likely to fail again at a later decision point from the same patient history.
Third, among different EHR access strategies, a medically motivated \textbf{active-search agent} performs best by first identifying the current clinical decision target and then actively gathering relevant patient-specific evidence before answering.

\begin{figure*}[t]
  \centering
  \includegraphics[width=0.9\textwidth, trim=150pt 0pt 70pt 0,clip]{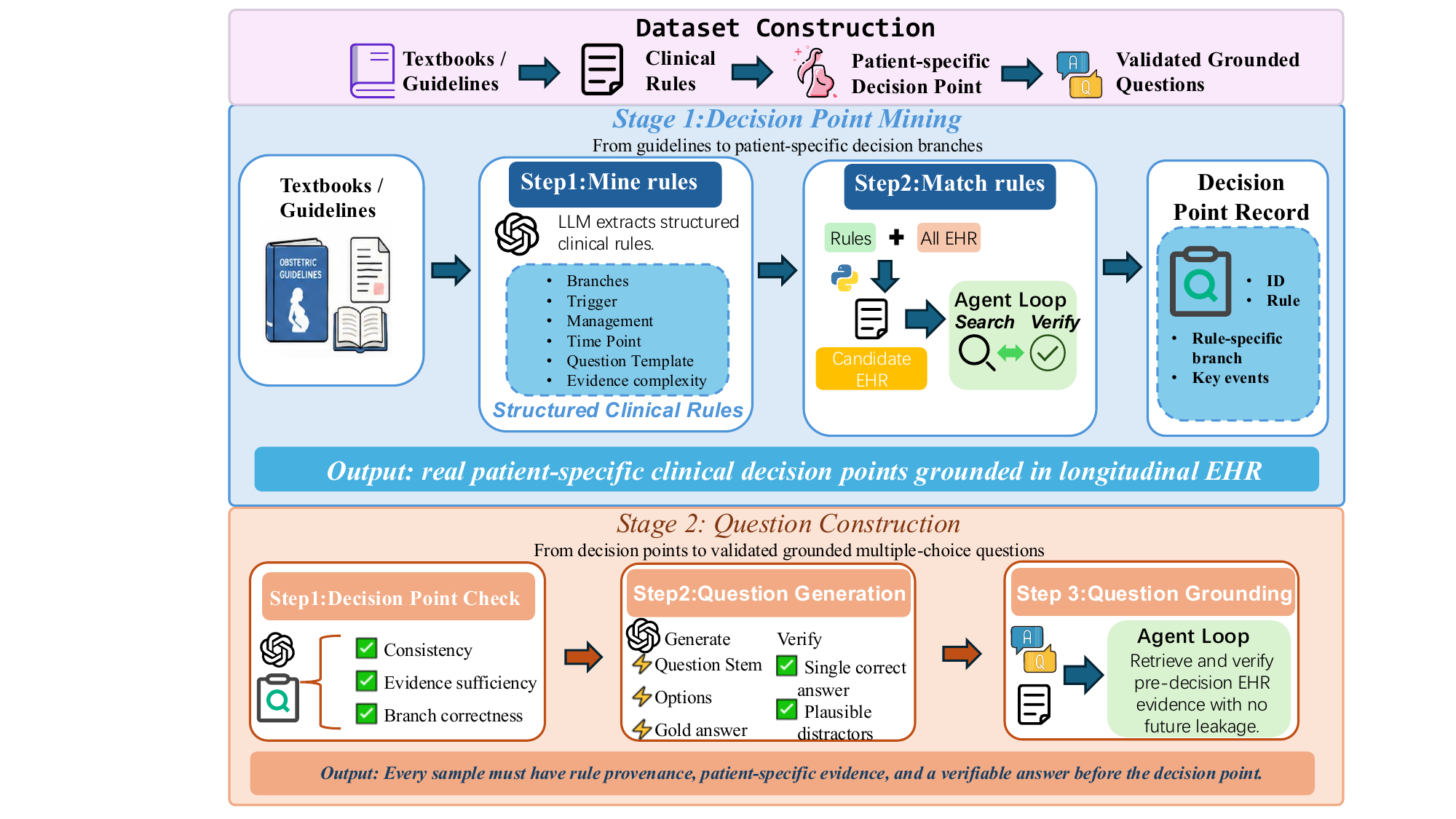}
\caption{Construction pipeline of ObGynLongBench. Stage 1 derives traceable clinical decision rules from obstetrics and gynecology textbooks and guidelines, structures them into executable rules, and matches them to longitudinal EHR histories through programmatic screening and agent-based review. Stage 2 constructs rule-grounded clinical questions, verifies patient-specific evidence, enforces the pre-decision cutoff, and generates the final benchmark cases.}
  \label{fig:construction-pipeline}
\end{figure*}

\section{ObGynLongBench}
\label{sec:benchmark}

Overall, we organize real pregnancy EHRs into longitudinal patient histories, derive traceable clinical decision rules from textbooks and guidelines, and use these rules to locate corresponding decision points within each patient's EHR history.
Each benchmark case is grounded in a real patient, a concrete gestational-timeline point, and a rule-defined clinical decision.

\subsection{Task Definition}
\label{sec:task-definition}

Each evaluation sample corresponds to a \textbf{clinical decision point} in a real patient's longitudinal EHR history.
At each decision point, we construct one sample consisting of a patient EHR input and a decision-oriented clinical question:
\[
c = (x, q),
\]
where $x$ denotes the patient EHR available at the decision point, and $q$ denotes the clinical question to be answered.

For a patient $p$, the longitudinal EHR history is represented as an ordered sequence of timestamped records:
\[
\begin{gathered}
\mathcal{H}_p =
\bigl\{(t_1, e_1), (t_2, e_2), \ldots, (t_n, e_n)\bigr\}, \\
t_1 < t_2 < \cdots < t_n .
\end{gathered}
\]
Each $e_i$ contains heterogeneous clinical events recorded at time $t_i$.
For a decision point at time $t_k$, the input EHR is the prefix of the longitudinal EHR history:
\[
x =
\mathcal{H}_{p,\leq k}
=
\bigl\{(t_1, e_1), (t_2, e_2), \ldots, (t_k, e_k)\bigr\}.
\]
All records after $t_k$ are removed to avoid future information leakage.

The question $q$ specifies the clinical decision to be made at $t_k$:
\[
q = (m, \mathcal{E}, r),
\]
where $m$ denotes the multiple-choice question instance, including the question stem, candidate answers, and the gold-standard answer;
$\mathcal{E} \subseteq x$ denotes the EHR evidence supporting the answer;
and $r$ denotes the clinical rule from which the decision point is derived.
Given only the pre-decision EHR prefix $x$, the LLM selects the correct option for $m$ based on the evidence contained in $x$.

\begin{figure*}[t]
  \centering
  \includegraphics[width=0.9\textwidth]{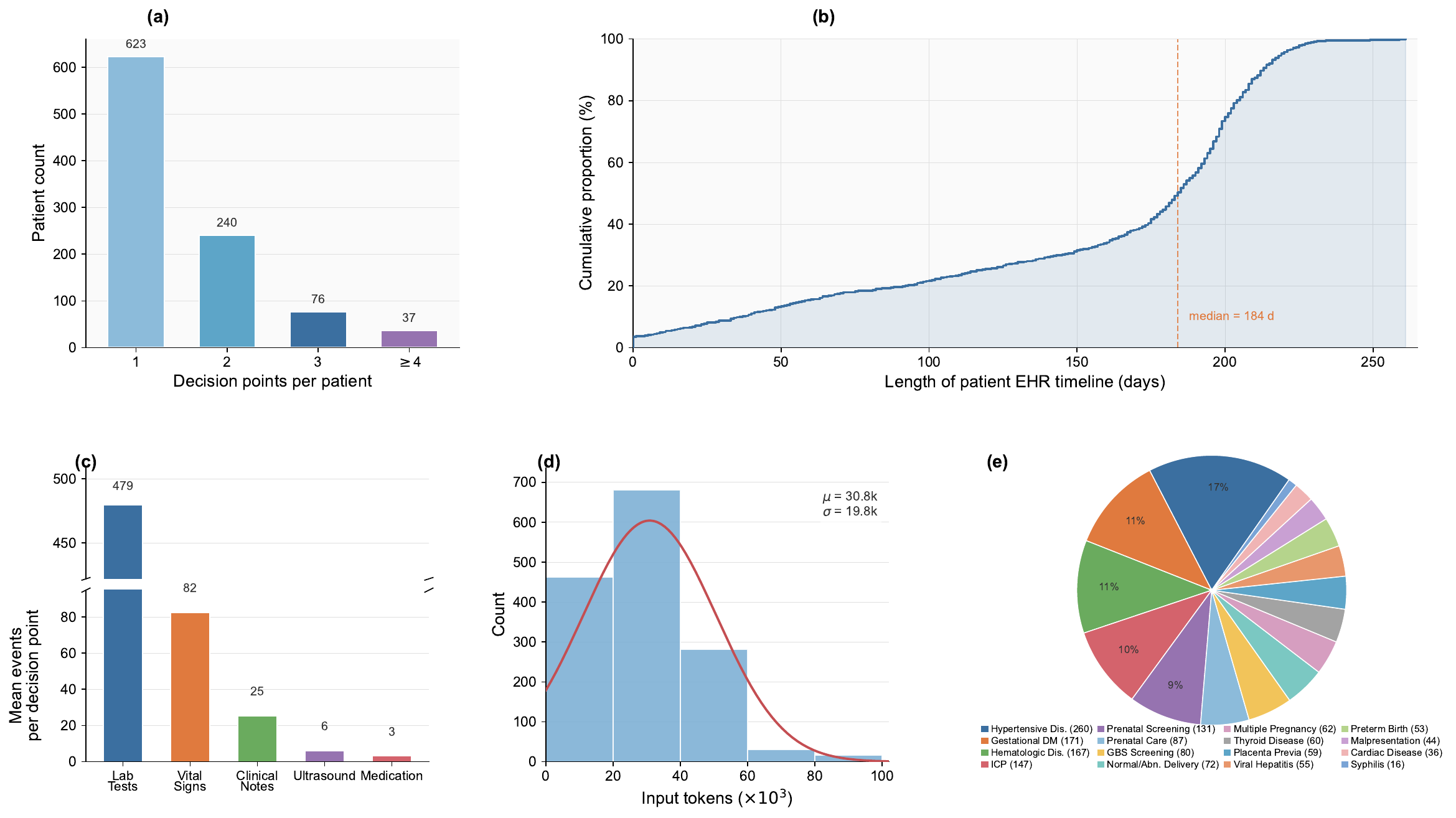}
\caption{Dataset overview of ObGynLongBench. \textbf{(a)} Distribution of decision points per patient. \textbf{(b)} Cumulative distribution of patient EHR history duration. \textbf{(c)} Mean number of pre-decision clinical events per decision point across five major event types. \textbf{(d)} Distribution of pre-decision input length under the \textbf{History-level EHR} setting, measured by the Qwen3.5 tokenizer. \textbf{(e)} Case distribution across 16 obstetric and gynecologic subspecialties.}
  \label{fig:dataset-overview}
\end{figure*}

\subsection{Data Construction}
\label{sec:data-construction}

ObGynLongBench is built from real pregnancy EHR histories and traceable obstetric and gynecologic clinical rules.
The EHRs come from a single-center tertiary hospital cohort, covering the pregnancy process from early registration to pre-delivery assessment.
Clinical rules are derived from textbooks and guidelines, define decision-point types, and guide question construction and evidence grounding.

Figure~\ref{fig:construction-pipeline} summarizes the construction pipeline.
Inspired by MIMIC-Instr~\citep{NEURIPS2024_62986e0a}, we align heterogeneous EHR events onto a gestational-age timeline, derive structured clinical rules, match them to longitudinal EHR histories through programmatic screening and agent-based review, and construct grounded decision-point questions.
The complete construction pipeline and independent clinician audit are presented in Appendix~\ref{app:data-construction}.

\subsection{Dataset}
\label{sec:dataset}

Figure~\ref{fig:dataset-overview} summarizes the scale and composition of the underlying EHR cohort and the resulting benchmark.
The dataset is built from 976 real pregnancy EHR histories and instantiated into 1,500 rule-grounded decision-point cases across 16 obstetric and gynecologic subspecialties.

Most patients contribute one decision point, while 240 patients contribute two and smaller groups contribute three or more.
The longitudinal EHR histories span a median of 184 days from early registration to pre-delivery assessment.
At each decision point, the available pre-decision EHR contains rich heterogeneous clinical events, with an average of 479 laboratory tests, 82 vital-sign measurements, 25 clinical notes, 6 ultrasound reports, and 3 medication records.
Under the \textbf{History-level EHR} setting, the mean pre-decision input length is 30.8K tokens ($\sigma=19.8$K).
The cases cover 16 obstetric and gynecologic subspecialties, including hypertensive disorders, gestational diabetes, hematologic disorders, ICP, prenatal screening, and other pregnancy-related conditions.

\section{Experiments}
\label{sec:experiments}

\begin{table*}
\centering
\footnotesize
\resizebox{0.9\textwidth}{!}{
\begin{tabular}{lc|c|cccc|c}
\hline
\rowcolor[HTML]{D9E2F3} 
\multicolumn{2}{c|}{\cellcolor[HTML]{D9E2F3}\textbf{Model}}                          & \textbf{Evidence-only} & \multicolumn{4}{c|}{\cellcolor[HTML]{D9E2F3}\textbf{History-level EHR}} & \textbf{Visit-level EHR} \\ \hline
\rowcolor[HTML]{D9E2F3} 
\multicolumn{1}{c|}{\cellcolor[HTML]{D9E2F3}\textbf{Name}} & \multicolumn{1}{c|}{\cellcolor[HTML]{D9E2F3}\textbf{Context window}} & \textbf{Overall}       & \textbf{Overall}    & \textbf{L1}    & \textbf{L2}    & \multicolumn{1}{c|}{\cellcolor[HTML]{D9E2F3}\textbf{L3}}    & \textbf{Overall}         \\ \hline
\rowcolor[HTML]{F2F2F2} 
\multicolumn{8}{c}{\cellcolor[HTML]{F2F2F2}\textbf{Commercial LLMs (Flash Level)}}                                                                                                                               \\ \hline
Claude-Haiku-4.5                                          & 256k                    & 75.5               & \accgaprow{59.8}{-15.7}     & 63.0           & 61.0           & 52.3           & \accgaprow{58.5}{1.3}            \\
GPT-5.4-mini                                              & 400k                    & 77.4               & \accgaprow{64.9}{-12.5}     & 61.8           & 68.0           & 60.7           & \accgaprow{62.3}{2.6}            \\
Gemini-3-Flash                                            & 1000k                   & 79.1               & \accgaprow{68.7}{-10.4}     & 66.0           & 71.5           & 65.0           & \accgaprow{67.1}{1.6}            \\
DeepSeek-V4-Flash                                         & 1000k                   & 78.8               & \accgaprow{64.4}{-14.4}     & 66.5           & 66.5           & 56.0           & \accgaprow{61.2}{3.2}            \\ \hline
\rowcolor[HTML]{F2F2F2} 
\multicolumn{8}{c}{\cellcolor[HTML]{F2F2F2}\textbf{Long-context open-source LLMs}}                                                                                                                               \\ \hline
Qwen3-VL-4B                                               & 262k                    & 72.1               & \accgaprow{59.8}{-12.3}     & 61.8           & 58.2           & 61.3           & \accgaprow{54.7}{5.1}            \\
Qwen3-VL-8B                                               & 262k                    & 74.3               & \accgaprow{61.8}{-12.5}     & 64.8           & 62.4           & 56.3           & \accgaprow{58.4}{3.4}            \\
Qwen3.5-4B                                                & 262k                    & 75.4               & \accgaprow{59.4}{-16.0}     & 63.2           & 57.5           & 59.3           & \accgaprow{52.7}{6.7}            \\
Qwen3.5-9B                                                & 262k                    & 73.1               & \accgaprow{59.3}{-13.8}     & 62.7           & 60.1           & 52.3           & \accgaprow{55.4}{3.9}            \\
Qwen3.5-35B-A3B                                           & 262k                    & 76.9               & \accgaprow{60.1}{-16.8}     & 58.0           & 62.1           & 57.7           & \accgaprow{55.7}{4.4}            \\
Gemma-4-E4B                                               & 131k                    & 68.1               & \accgaprow{52.0}{-16.1}     & 56.0           & 53.2           & 43.3           & \accgaprow{47.4}{4.6}            \\
Gemma-4-26B-A4B                                           & 262k                    & 78.2               & \accgaprow{60.7}{-17.5}     & 65.0           & 61.4           & 53.0           & \accgaprow{60.6}{0.1}            \\
Phi-4-mini                                                & 131k                    & 58.7               & \accgaprow{51.1}{-7.6}      & 56.2           & 47.8           & 53.3           & \accgaprow{48.6}{2.5}            \\ \hline
\rowcolor[HTML]{F2F2F2} 
\multicolumn{8}{c}{\cellcolor[HTML]{F2F2F2}\textbf{Long-context Medical LLMs}}                                                                                                                                   \\ \hline
HealthGPT-Pro-4B                                          & 262k                    & 69.6               & \accgaprow{57.1}{-12.5}     & 59.0           & 56.1           & 57.3           & \accgaprow{54.1}{3.0}            \\
HealthGPT-Pro-8B                                          & 262k                    & 73.5               & \accgaprow{61.3}{-12.2}     & 67.2           & 59.8           & 57.3           & \accgaprow{58.1}{3.2}            \\
Hulu-Med-30A3                                             & 75k                     & 74.9               & \accgaprow{53.9}{-21.0}     & 59.2           & 53.0           & 49.0           & \accgaprow{55.8}{-1.9}           \\
Lingshu-7B                                                & 128k                    & 72.1               & \accgaprow{59.6}{-12.5}     & 61.0           & 59.5           & 58.0           & \accgaprow{60.4}{-0.8}           \\
MedGemma-1.5-4B                                           & 131k                    & 64.0               & \accgaprow{41.3}{-22.7}     & 45.8           & 43.5           & 29.7           & \accgaprow{48.1}{-6.8}          
\end{tabular}
}
\caption{Main results on ObGynLongBench. We report accuracy (\%) for each LLM under three input-scope settings: \textbf{Evidence-only}, \textbf{History-level EHR}, and \textbf{Visit-level EHR}. Parenthesized values in the \textbf{Overall} column of \textbf{History-level EHR} denote the accuracy gap relative to \textbf{Evidence-only} ($\Delta = \textit{History} - \textit{Evidence}$), and those in the \textbf{Overall} column of \textbf{Visit-level EHR} denote the gap relative to \textbf{History-level EHR} ($\Delta = \textit{History} - \textit{Visit}$); positive gaps are shown in green as $(+\Delta)$ and negative gaps in red as $(-\Delta)$. For open-source and medical LLMs, thinking mode is disabled for fair comparison because it often causes context-window overflow.}
\label{tab:scope}
\end{table*}

This section builds on the benchmark and evaluation protocol defined in Section~\ref{sec:benchmark}.
We first describe the experimental setup and then report the main results under the three input-scope settings.

\subsection{Experimental Setup}
\label{sec:models-setup}

\paragraph{Information Scope.}
For each case $c=(x,q)$ with $q=(m,\mathcal{E},r)$, we evaluate LLMs under three input-scope settings at the same decision point $t_k$:
\textbf{Evidence-only} (\textbf{Evidence}) provides only the supporting evidence $\mathcal{E} \subseteq x$;
\textbf{Visit-level EHR} (\textbf{Visit}) provides all EHR records recorded on the same day as the decision point $t_k$;
and \textbf{History-level EHR} (\textbf{History}) provides the full pre-decision EHR history $x=\mathcal{H}_{p,\leq k}$.
The LLM is required to answer $m$ from the provided input under each setting.

\paragraph{Models.}
We evaluate 17 LLMs from three groups that are practically relevant to personalized medical assistant scenarios:
\textbf{Commercial LLMs}, which provide fast and accessible inference, including Claude Haiku 4.5~\citep{Anthropic2025ClaudeHaiku45}, GPT-5.4-mini~\citep{OpenAI2026GPT54MiniNano}, Gemini-3-Flash~\citep{Google2025Gemini3Flash}, and DeepSeek-V4-Flash~\citep{deepseekai2026deepseekv4};
\textbf{Long-context open-source LLMs}, including Qwen3-VL~\citep{bai2025qwen3vltechnicalreport}, Qwen3.5~\citep{qwen3.5},  Gemma 4~\citep{GoogleDeepMind2026Gemma4ModelCard}, and the Phi-4 series~\citep{microsoft2025phi4minitechnicalreportcompact};
and \textbf{Long-context medical LLMs}, including HealthGPT-Pro~\citep{pmlr-v267-lin25n}, Hulu-Med~\citep{jiang2025hulumedtransparentgeneralistmodel}, Lingshu~\citep{xu2025lingshu}, and MedGemma-1.5-4B~\citep{sellergren2026medgemma}.

\paragraph{Evaluation metric.}
All experiments use a multiple-choice question (MCQ) format, where each LLM outputs a single answer option.
Given a test set $\mathcal{D}$, accuracy is defined as:
\begin{equation}
\label{eq:acc}
\mathrm{Acc}
=
\frac{1}{|\mathcal{D}|}
\sum_{i=1}^{|\mathcal{D}|}
\mathbb{I}(\hat{y}_i=y_i),
\end{equation}
where $\hat{y}_i$ is the LLM-predicted answer and $y_i$ is the gold-standard answer.

\subsection{Main Results}
\label{sec:main-results}

Table~\ref{tab:scope} reports the main results of the three LLM groups under three input-scope settings:
\textbf{Evidence}, \textbf{Visit}, and \textbf{History}, corresponding to \textbf{Evidence-only}, \textbf{Visit-level EHR}, and \textbf{History-level EHR}, respectively.
Patient-clustered bootstrap confidence intervals are reported in Appendix~\ref{app:bootstrap-uncertainty}.

\paragraph{The gap between Evidence-only and History-level EHR is large.}
Under \textbf{Evidence}, the best-performing LLMs in the three groups are Gemini-3-Flash (79.1\%), Gemma-4-26B-A4B (78.2\%), and Hulu-Med-30A3 (74.9\%).
Under \textbf{History}, the best scores drop to Gemini-3-Flash (68.7\%), Qwen3-VL-8B (61.8\%), and HealthGPT-Pro-8B (61.3\%).
This shows that \textbf{Evidence-only} performance does not directly transfer to \textbf{History-level EHR}.

\paragraph{Medical LLMs show a larger gap under History-level EHR.}
Medical LLMs do not show a clear advantage in the \textbf{History} setting.
The gap between the best general-purpose LLM and the best medical LLM increases from 4.2pp under \textbf{Evidence} to 7.4pp under \textbf{History}.
This suggests that existing medical LLM training does not sufficiently prepare LLMs to extract decision-relevant information from realistic EHR histories.

\paragraph{Visit is incomplete; History can introduce additional noise.}
Under \textbf{Visit}, LLM performance drops substantially compared with \textbf{Evidence}, indicating that same-day records often lack the longitudinal context needed for obstetric and gynecologic decision-making.
For most LLMs, \textbf{History} improves over \textbf{Visit}, showing that longitudinal EHR history provides useful clinical context.
However, for LLMs with context windows shorter than 256K, \textbf{History} performs worse than \textbf{Visit}.
This indicates that full history-level EHR input can also introduce additional noise, especially when LLMs cannot reliably process long EHRs.

\section{Further Analysis}
\label{sec:further-analysis}

Section~\ref{sec:main-results} establishes the \textbf{Evidence-to-EHR Gap} under different input scopes.
This section provides additional analyses under the \textbf{History-level EHR} setting, examining how evidence complexity, EHR input length, and patient-level history affect LLM performance.

\subsection{Evidence Complexity Analysis}
\label{sec:evidence-complexity}

Under the \textbf{History} setting, we further analyze how LLM performance changes with evidence complexity.
Clinical rules are assigned to three evidence-complexity levels, and cases are grouped by the rule used to construct them:
\textbf{L1 (single-evidence recognition)} depends on one explicit and locally visible key evidence item;
\textbf{L2 (multi-source integration)} requires combining multiple evidence items or modalities within the same visit or local context;
and \textbf{L3 (long-horizon reasoning)} requires retrieving, time-anchoring, and integrating evidence distributed across the longitudinal EHR history.
Full stratification criteria are provided in Appendix~\ref{sec:difficulty-metrics}.

Figure~\ref{fig:accuracy-by-mode} shows a consistent ordering under the \textbf{History} setting: accuracy is highest on L1, lower on L2, and lowest on L3.
This indicates that the difficulty of locating and integrating the supporting evidence $\mathcal{E} \subseteq x$ from the pre-decision EHR history directly affects LLM performance under the \textbf{History} setting.

\begin{figure}[t]
  \centering
  \includegraphics[width=\columnwidth]{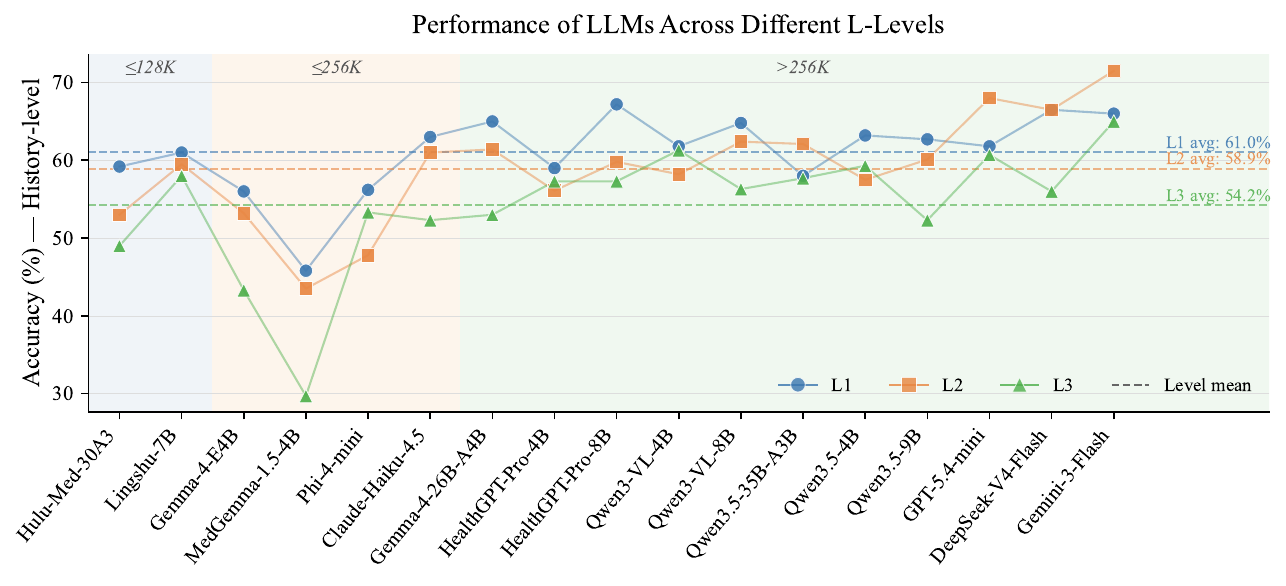}
  \caption{LLM performance across evidence-complexity levels under the \textbf{History-level EHR} setting. Accuracy decreases from L1 to L3 as evidence utilization and integration become more difficult.}
  \label{fig:accuracy-by-mode}
\end{figure}

\begin{figure}[t]
  \centering
  \includegraphics[width=\columnwidth]{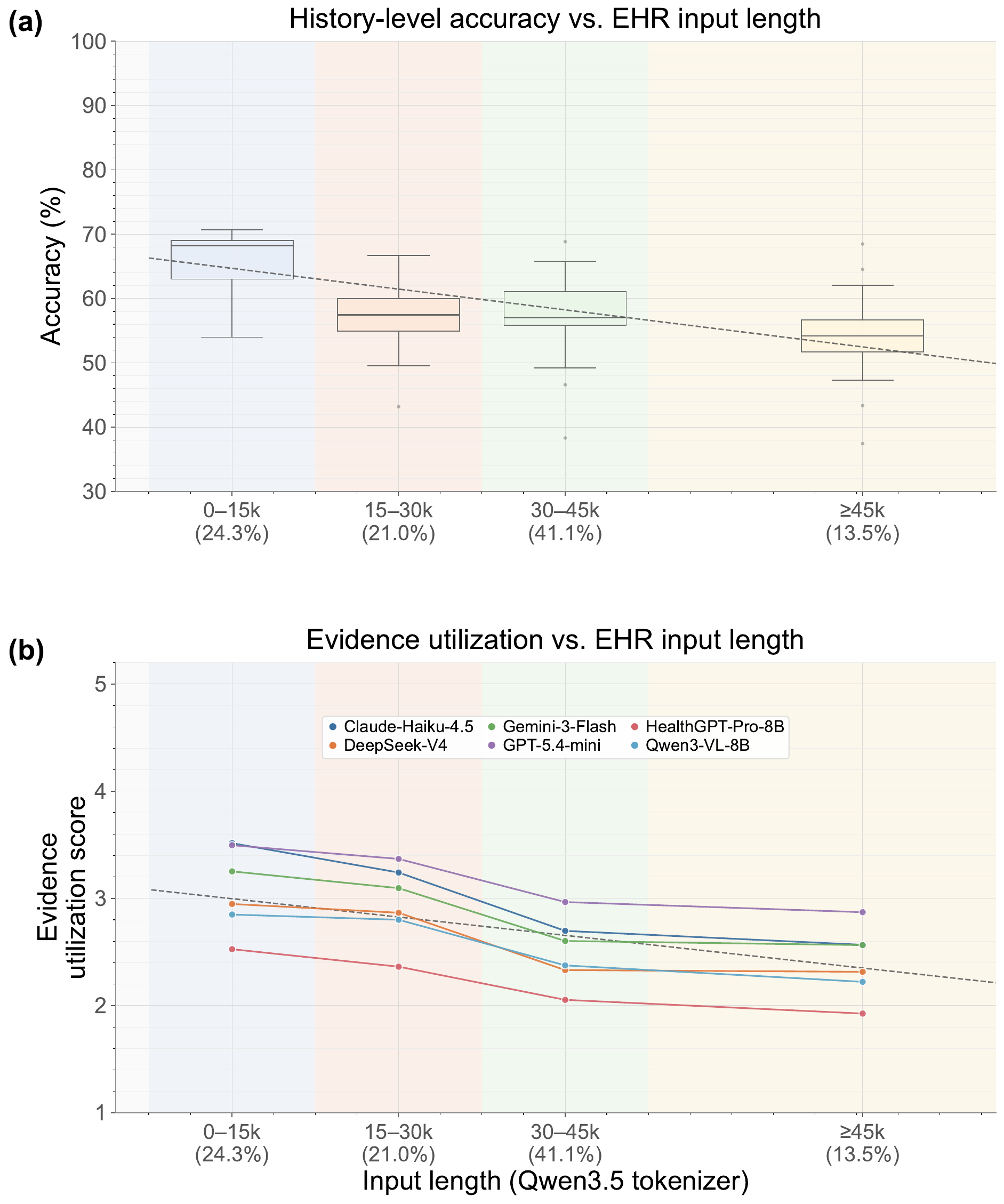}
\caption{Context-length analysis under \textbf{History-level EHR}. \textbf{(a)} Accuracy across pre-decision EHR input-length bins, with boxes computed over 17 LLMs. \textbf{(b)} LLM-judged evidence utilization across the same bins for six representative LLMs.}
  \label{fig:evidence-accuracy-input}
\end{figure}

\subsection{Context Length Analysis}
\label{sec:context-length}

We further examine whether longer EHR histories make the \textbf{History} setting more difficult.
Specifically, we group cases by the length of the pre-decision EHR input and analyze both final accuracy and evidence utilization.

\paragraph{Longer EHR histories lead to lower accuracy.}
Figure~\ref{fig:evidence-accuracy-input}(a) reports LLM accuracy under the \textbf{History} setting across four input-length bins.
Accuracy consistently decreases as the pre-decision EHR input becomes longer.

\paragraph{The decline is accompanied by weaker evidence utilization.}
To understand why longer EHR inputs reduce accuracy, we use DeepSeek-V4-Flash as the primary LLM judge to assess evidence utilization by comparing each LLM's rationale with the gold supporting evidence $\mathcal{E}$ (Appendix~\ref{app:judge-robustness}).
Figure~\ref{fig:evidence-accuracy-input}(b) shows that evidence utilization also decreases as input length increases across six representative LLMs.
This suggests that the accuracy drop is partly driven by weaker evidence utilization: as the EHR history becomes longer, LLMs are less able to identify and use decision-relevant evidence, leading to less reliable clinical decisions.

\subsection{Patient-level Analysis}
\label{sec:patient-stability}

We further analyze 240 patients who each have two consecutive decision points in the same pregnancy EHR history.
For each patient, we denote the two decision points as DP1 and DP2.
To control for gestational timing, we compare DP2 with a temporally matched control group: DP1 cases after 30 gestational weeks from patients with only one decision point.
Table~\ref{tab:retention-gap} reports four quantities under \textbf{History-level EHR}: overall DP2 accuracy, accuracy on the temporally matched DP1 control group, DP2 accuracy when DP1 is answered incorrectly, and DP2 accuracy when DP1 is answered correctly.
Values in parentheses denote the change relative to overall DP2 accuracy.

At the aggregate level, LLM performance on DP2 is close to that on the temporally matched DP1 control group, suggesting that DP2 has comparable difficulty to other decision points from the same pregnancy period.
However, a clear patient-level pattern emerges when we condition on whether the earlier decision point from the same patient is answered correctly.
For most LLMs, DP2 accuracy decreases when DP1 is wrong and increases when DP1 is correct.
This indicates that LLM errors are patient-dependent: if an LLM fails to interpret one decision point in a patient's EHR history, it is more likely to fail again on a later decision point from the same history.
In realistic personalized medical assistant settings, LLMs would often be used repeatedly across multiple decision points throughout the same patient's longitudinal care.
Appendix~\ref{app:expanded-patient-analysis} further examines patient-level performance using all 1,500 cases and all 17 models.

\begin{table}[htbp]
  \centering
  \footnotesize
  \setlength{\tabcolsep}{1pt}
  \resizebox{\columnwidth}{!}{%
  \begin{tabular}{lcccc}
    \toprule
    \textbf{Model} &
    \textbf{\begin{tabular}[c]{@{}c@{}}Acc@DP2\\(All)\end{tabular}} &
    \textbf{\begin{tabular}[c]{@{}c@{}}Acc@DP1\\($GA \geq 30$w)\end{tabular}} &
    \textbf{\begin{tabular}[c]{@{}c@{}}Acc@DP2\\(DP1 wrong)\end{tabular}} &
    \textbf{\begin{tabular}[c]{@{}c@{}}Acc@DP2\\(DP1 correct)\end{tabular}} \\
    \midrule
    Gemini-3-Flash      & 67.1            & \accgaprow{69.3}{2.2}   & \accgaprow{66.7}{-0.4}   & \accgaprow{67.3}{0.2}   \\
    GPT-5.4-mini        & 65.4            & \accgaprow{63.3}{-2.1}   & \accgaprow{64.4}{-1.0}   & \accgaprow{65.9}{0.5}   \\
    DeepSeek-V4-Flash   & 61.7            & \accgaprow{64.0}{2.3}   & \accgaprow{59.0}{-2.7}   & \accgaprow{63.0}{1.3}   \\
    HealthGPT-Pro-8B    & 57.5            & \accgaprow{59.5}{2.0}   & \accgaprow{53.3}{-4.2}   & \accgaprow{60.1}{2.6}   \\
    Claude-Haiku-4.5    & 57.1            & \accgaprow{57.1}{0.0}   & \accgaprow{48.8}{-8.3}   & \accgaprow{61.4}{4.3}   \\
    Qwen3-VL-8B         & 57.1            & \accgaprow{57.6}{0.5}   & \accgaprow{51.2}{-5.9}   & \accgaprow{60.3}{3.2}   \\
    \bottomrule
  \end{tabular}%
  }
\caption{Patient-level analysis under \textbf{History-level EHR}. We report accuracy (\%). Values in parentheses denote changes relative to overall DP2 accuracy.}
  \label{tab:retention-gap}
\end{table}

\section{EHR Access Strategies under History-level EHR}
\label{sec:ehr-access}

Section~\ref{sec:context-length} shows that direct reading of the full pre-decision EHR history cannot recover the performance achieved under the \textbf{Evidence} setting.
We therefore compare alternative EHR access strategies under the \textbf{History} input scope.
For each case, we keep the same full pre-decision EHR history fixed and vary only how LLMs access, organize, and compress the records.
Our goal is to identify strategies that improve accuracy while controlling token cost in long-context medical record scenarios.

\subsection{Strategy Settings}
\label{sec:strategy-settings}

We use \textbf{Direct} access to the \textbf{History-level EHR} input as the baseline.
We compare five alternative EHR access strategies against this baseline under the same full pre-decision EHR history, varying only how LLMs represent, retrieve, and interact with the records.

\paragraph{Image.}
We render the full pre-decision EHR history into multiple page-level images and provide them to LLMs.

\paragraph{Static RAG.}
We split the EHR history into date-based chunks, retrieve relevant snippets according to the current question in a single step, and provide them to LLMs.

\paragraph{Static RAG + Recency (Static RAG + Rec.).}
We add recency weighting to Static RAG, assigning higher retrieval weights to records closer to the decision point.

\paragraph{Rolling Summary.}
We input the EHR history every 7 days in chronological order and ask LLMs to generate stage-wise summaries.
Each subsequent round contains the previous summary and the new 7-day records.

\paragraph{Agent.}
The agent first reads the clinical question to identify the decision target, then obtains basic patient information to anchor the case in the EHR history.
It then actively searches for decision-relevant evidence through an iterative retrieval loop, rather than relying on a single-pass reading of the full \textbf{History-level EHR} input.

Implementation details for the RAG and Agent settings are provided in Appendix~\ref{app:implementation-details}.

\subsection{Strategy Comparison}
\label{sec:strategy-results}

Table~\ref{tab:qwen-modes} compares the accuracy and average token cost of each EHR access strategy, averaged over three Qwen3.5-series LLMs under the \textbf{History-level EHR} setting.
Values in parentheses indicate changes relative to the \textbf{Direct} baseline.
Paired-bootstrap uncertainty estimates for the overall accuracy differences relative to \textbf{Direct} are provided in Appendix~\ref{app:access-bootstrap}.
For \textbf{Rolling Summary}, token cost is computed cumulatively over the full rolling summarization process, rather than using only the input length of the final round.

\begin{table}[htbp]
  \centering
  \footnotesize
  \setlength{\tabcolsep}{2pt}
  \resizebox{\columnwidth}{!}{%
  \begin{tabular}{lcccc|r}
    \toprule
    \textbf{Strategy} & \textbf{Overall} & \textbf{L1} & \textbf{L2} & \textbf{L3} & \textbf{Tok.} \\
    \midrule
    Direct             & 59.6            & 61.3            & 59.9            & 56.4            & 30,951 \\
    Image              & \accgaprow{52.7}{-6.9}   & \accgaprow{55.1}{-6.2}   & \accgaprow{52.8}{-7.1}   & \accgaprow{49.0}{-7.4}   & 11,433 \\
    Static RAG         & \accgaprow{58.9}{-0.7}   & \accgaprow{58.8}{-2.5}   & \accgaprow{59.7}{-0.2}   & \accgaprow{56.8}{0.4}   &  4,308 \\
    Static RAG + Rec.  & \accgaprow{59.2}{-0.4}   & \accgaprow{60.6}{-0.7}   & \accgaprow{60.1}{0.2}   & \accgaprow{54.5}{-1.9}   &  4,089 \\
    Rolling Summary    & \accgaprow{57.9}{-1.7}   & \accgaprow{62.4}{1.1}   & \accgaprow{58.1}{-1.8}   & \accgaprow{51.7}{-4.8}   & 61,968 \\
    Agent              & \accgaprow{64.2}{4.6}   & \accgaprow{63.8}{2.5}   & \accgaprow{65.1}{5.2}   & \accgaprow{62.1}{5.7}   & 35,786 \\
    \bottomrule
  \end{tabular}%
  }
\caption{Accuracy (\%) and average token cost of different EHR access strategies under the \textbf{History-level EHR} setting, averaged over three Qwen3.5-series LLMs. Values in parentheses denote changes relative to the \textbf{Direct} baseline.}
  \label{tab:qwen-modes}
\end{table}

\paragraph{Agent.}
\textbf{Agent} achieves the best overall accuracy (64.2\%, +4.6pp) and improves over \textbf{Direct} access across all evidence complexity levels, with the largest gain on L3 (+5.7pp).
This suggests that history-level EHR reasoning benefits from question-guided active search: LLMs should first identify the current clinical decision target and then actively locate decision-relevant evidence, rather than simply shorten or read the full input in a single pass.

\paragraph{Static RAG and Static RAG + Rec.}
Both retrieval-based strategies substantially reduce token cost while maintaining accuracy close to \textbf{Direct} access.
\textbf{Static RAG} performs slightly better on L2/L3, whereas \textbf{Static RAG + Rec.} improves L1 and L2 but drops on L3.
This indicates a recency bias: emphasizing recent records helps locally anchored decisions, but may miss earlier evidence needed for long-horizon reasoning.

\paragraph{Image.}
\textbf{Image} reduces token count but degrades performance at all complexity levels, especially on L3 ($-$7.4pp).
This suggests that page-level rendering introduces visual parsing noise and hinders cross-page evidence integration.

\paragraph{Rolling Summary.}
\textbf{Rolling Summary} incurs the highest token cost and does not surpass \textbf{Direct} overall ($-$1.7pp).
Its drop on L3 ($-$4.8pp) suggests that sequential summarization may lose cross-time evidence needed for harder questions.

\section{Related Work}
\label{sec:related}

\paragraph{Medical and obstetric QA benchmarks.}
Medical LLM evaluation commonly relies on examinations and benchmarks, including MedQA~\citep{app11146421}, MedMCQA~\citep{pmlr-v174-pal22a}, PubMedQA~\citep{jin-etal-2019-pubmedqa}, MultiMedQA~\citep{Singhal2022LargeLM}, and MedRCube~\citep{bao2026medrcubemultidimensionalframeworkfinegrained}. In static QA benchmarks, relevant clinical evidence is usually provided in the question.
Obstetric and gynecologic evaluations similarly focus on specialty examinations, board-style questions, or patient-facing maternal health QA~\citep{Bachmann2024ExploringTC,ugowski2025ComparativeAO,srikanth-etal-2024-pregnant}, evaluating domain knowledge and response quality rather than reasoning over real longitudinal pregnancy EHR histories.

\paragraph{EHR benchmarks.}
Recent EHR benchmarks move closer to real clinical records.
MedAlign evaluates clinician-generated instructions grounded in longitudinal EHRs~\citep{Fleming2023MedAlignAC};
MIMIC-Instr constructs large-scale EHR-grounded instruction-following data from MIMIC-IV~\citep{NEURIPS2024_62986e0a};
and TIMER focuses on temporal reasoning over longitudinal patient histories~\citep{Cui2025TIMERTI}.
These benchmarks reveal that EHR-based reasoning is more difficult than static medical QA.
However, they mainly target general EHR tasks such as retrieval, summarization, instruction following, or temporal reasoning.
In contrast, ObGynLongBench focuses on rule-grounded obstetric decision points along the pregnancy EHR history, requiring LLMs to retrieve, time-anchor, and integrate patient-specific evidence before making clinical decisions.

\section{Conclusion}
\label{sec:conclusion}

We introduced \textbf{ObGynLongBench}, a rule-grounded long-context EHR benchmark for obstetric and gynecologic decision-making.
Experiments reveal a clear \textbf{Evidence-to-EHR Gap}: LLMs perform well with directly provided evidence, but accuracy drops under \textbf{Visit-level EHR} and \textbf{History-level EHR}.
Further analyses show that this gap is driven by evidence utilization and worsens with longer and more complex EHR evidence; model errors are also associated across longitudinal decision points within patients.
Active-search agents perform best among EHR access strategies, highlighting patient-specific evidence utilization as a key requirement for reliable personalized medical assistants.

\section*{Limitations}

This paper has the following limitations:

\begin{itemize}
\item \textbf{Single-center EHR cohort.}
ObGynLongBench is built from real pregnancy EHRs from a single tertiary hospital, and may reflect institution-specific documentation habits, clinical workflows, and regional population characteristics.
Future work should validate the benchmark across hospitals and regions.

\item \textbf{Chinese-language clinical records.}
The EHRs are written in Chinese, so performance may be affected by Chinese clinical language understanding, abbreviation handling, and multilingual robustness.
The conclusions may not directly generalize to EHRs in other languages.

\item \textbf{MCQ-based evaluation.}
We use MCQs for controlled automatic evaluation, but real clinical assistance often requires open-ended reasoning, structured recommendations, uncertainty expression, and interactive clarification.
Future benchmarks should extend to more realistic clinical output formats.

\item \textbf{Inference-setting differences.}
As detailed in Appendix~\ref{app:implementation-details}, open-source and medical LLMs are evaluated with thinking mode disabled, whereas commercial LLMs use their APIs' default inference configurations.
These differences may affect absolute performance comparisons across model families.
However, the \textbf{Evidence-to-EHR Gap} is measured within each model under a fixed configuration and therefore does not require identical inference settings across families.
\end{itemize}

\section*{Ethical Considerations}

This work uses real-world pregnancy EHRs, which contain sensitive health information.
All data processing and benchmark construction was conducted under institutional ethics approval and appropriate data-use agreements.
Patient records were used only in de-identified form, and the benchmark is intended solely for research evaluation.
Researchers should not attempt patient re-identification, linkage with external data sources, or use the benchmark to infer private attributes.
Consent requirements were handled according to the approved institutional protocol.

ObGynLongBench is not a clinical decision-support system.
Model outputs may be incomplete, poorly grounded, or clinically unsafe, even when benchmark accuracy is high.
Any real clinical use of LLMs would require careful validation, clinician oversight, patient consent where appropriate, and compliance with local medical, legal, and regulatory requirements.
The goal of this benchmark is to support safer research on long-context EHR reasoning, not to encourage autonomous medical deployment.

\section*{Acknowledgments}
This work was supported by the National Key R\&D Program of China (Grant No.~2023YFF1204800) and the AI for Science Program of the Shanghai Municipal Commission of Economy and Informatization (Grant No.~2025-GZL-RGZN-BTBX-02028).
This work was also supported by the National Natural Science Foundation of China (Grant No.~62406121) and the Shanghai Basic Research Program (Natural Science Foundation) (Grant No.~25ZR1401034).
Computational resources for this project were partially provided by the CFFF platform at Fudan University.

AI writing assistants were used only for language polishing and editing. All scientific claims, experimental results, analyses, and final manuscript content were reviewed and verified by the authors.

\bibliography{custom}

@article{Peahl2020TheEO,
  title={The evolution of prenatal care delivery guidelines in the United States},
  author={Peahl, Alex F. and Howell, Joel D.},
  journal={American Journal of Obstetrics and Gynecology},
  year={2021},
  volume={224},
  number={4},
  pages={339--347},
  doi={10.1016/j.ajog.2020.12.016}
}

@article{Lee2020SocietyFM,
  title={Society for Maternal-Fetal Medicine Consult Series \#53: Intrahepatic cholestasis of pregnancy: Replaces Consult \#13, April 2011},
  author={{Society for Maternal-Fetal Medicine (SMFM)} and Lee, Richard H. and Greenberg, Mara and Metz, Torri D. and Pettker, Christian M.},
  journal={American Journal of Obstetrics and Gynecology},
  year={2021},
  volume={224},
  number={2},
  pages={B2--B9},
  doi={10.1016/j.ajog.2020.11.002}
}

@article{2020PreventionOG,
  title={Prevention of Group B Streptococcal Early-Onset Disease in Newborns: ACOG Committee Opinion, Number 797.},
  author={{American College of Obstetricians and Gynecologists}},
  journal={Obstetrics \& Gynecology},
  year={2020},
  volume={135},
  number={2},
  pages={e51--e72},
  doi={10.1097/AOG.0000000000003668}
}

@article{2020GestationalHA,
  title={Gestational Hypertension and Preeclampsia: ACOG Practice Bulletin, Number 222.},
  author={{American College of Obstetricians and Gynecologists}},
  journal={Obstetrics \& Gynecology},
  year={2020},
  volume={135},
  number={6},
  pages={e237--e260},
  doi={10.1097/AOG.0000000000003891}
}

@article{Rose2020ScreeningFF,
  title={Screening for Fetal Chromosomal Abnormalities: ACOG Practice Bulletin, Number 226.},
  author={{American College of Obstetricians and Gynecologists' Committee on Practice Bulletins--Obstetrics} and {Committee on Genetics} and {Society for Maternal-Fetal Medicine}},
  journal={Obstetrics \& Gynecology},
  year={2020},
  volume={136},
  number={4},
  pages={e48--e69},
  doi={10.1097/AOG.0000000000004084}
}

@Article{app11146421,
AUTHOR = {Jin, Di and Pan, Eileen and Oufattole, Nassim and Weng, Wei-Hung and Fang, Hanyi and Szolovits, Peter},
TITLE = {What Disease Does This Patient Have? A Large-Scale Open Domain Question Answering Dataset from Medical Exams},
JOURNAL = {Applied Sciences},
VOLUME = {11},
YEAR = {2021},
NUMBER = {14},
ARTICLE-NUMBER = {6421},
URL = {https://www.mdpi.com/2076-3417/11/14/6421},
ISSN = {2076-3417},
DOI = {10.3390/app11146421}
}

@InProceedings{pmlr-v174-pal22a,
  title =   {MedMCQA: A Large-scale Multi-Subject Multi-Choice Dataset for Medical domain Question Answering},
  author =       {Pal, Ankit and Umapathi, Logesh Kumar and Sankarasubbu, Malaikannan},
  booktitle =   {Proceedings of the Conference on Health, Inference, and Learning},
  pages =   {248--260},
  year =   {2022},
  editor =   {Flores, Gerardo and Chen, George H and Pollard, Tom and Ho, Joyce C and Naumann, Tristan},
  volume =   {174},
  series =   {Proceedings of Machine Learning Research},
  month =   {07--08 Apr},
  publisher =    {PMLR},
  url =   {https://proceedings.mlr.press/v174/pal22a.html}
}

@inproceedings{jin-etal-2019-pubmedqa,
    title = "{P}ub{M}ed{QA}: A Dataset for Biomedical Research Question Answering",
    author = "Jin, Qiao  and
      Dhingra, Bhuwan  and
      Liu, Zhengping  and
      Cohen, William  and
      Lu, Xinghua",
    editor = "Inui, Kentaro  and
      Jiang, Jing  and
      Ng, Vincent  and
      Wan, Xiaojun",
    booktitle = "Proceedings of the 2019 Conference on Empirical Methods in Natural Language Processing and the 9th International Joint Conference on Natural Language Processing (EMNLP-IJCNLP)",
    month = nov,
    year = "2019",
    address = "Hong Kong, China",
    publisher = "Association for Computational Linguistics",
    url = "https://aclanthology.org/D19-1259/",
    doi = "10.18653/v1/D19-1259",
    pages = "2567--2577"
}

@article{Singhal2022LargeLM,
  title={Large language models encode clinical knowledge},
  author={Singhal, Karan and Azizi, Shekoofeh and Tu, Tao and Mahdavi, S. Sara and Wei, Jason and Chung, Hyung Won and Scales, Nathan and Tanwani, Ajay and Cole-Lewis, Heather and Pfohl, Stephen and Payne, Perry and Seneviratne, Martin and Gamble, Paul and Kelly, Chris and Babiker, Abubakr and Sch{\"a}rli, Nathanael and Chowdhery, Aakanksha and Mansfield, Philip and Demner-Fushman, Dina and Ag{\"u}era y Arcas, Blaise and Webster, Dale and Corrado, Greg S. and Matias, Yossi and Chou, Katherine and Gottweis, Juraj and Tomasev, Nenad and Liu, Yun and Rajkomar, Alvin and Barral, Joelle and Semturs, Christopher and Karthikesalingam, Alan and Natarajan, Vivek},
  journal={Nature},
  year={2023},
  volume={620},
  pages={172--180},
  doi={10.1038/s41586-023-06291-2}
}

@article{Goldberg2017CommitteeON,
  title={Committee Opinion No 700: Methods for Estimating the Due Date.},
  author={{American College of Obstetricians and Gynecologists}},
  journal={Obstetrics \& Gynecology},
  year={2017},
  volume={129},
  number={5},
  pages={e150--e154},
  doi={10.1097/AOG.0000000000002046}
}

@article{Fleming2023MedAlignAC,
  title={MedAlign: A Clinician-Generated Dataset for Instruction Following with Electronic Medical Records},
  author={Fleming, Scott L. and Lozano, Alejandro and Haberkorn, William J. and Jindal, Jenelle A. and Reis, Eduardo and Thapa, Rahul and Blankemeier, Louis and Genkins, Julian Z. and Steinberg, Ethan and Nayak, Ashwin and Patel, Birju and Chiang, Chia-Chun and Callahan, Alison and Huo, Zepeng and Gatidis, Sergios and Adams, Scott and Fayanju, Oluseyi and Shah, Shreya J. and Savage, Thomas and Goh, Ethan and Chaudhari, Akshay S. and Aghaeepour, Nima and Sharp, Christopher and Pfeffer, Michael A. and Liang, Percy and Chen, Jonathan H. and Morse, Keith E. and Brunskill, Emma P. and Fries, Jason A. and Shah, Nigam H.},
  journal={Proceedings of the AAAI Conference on Artificial Intelligence},
  year={2024},
  volume={38},
  number={20},
  pages={22021--22030},
  doi={10.1609/aaai.v38i20.30205}
}

@inproceedings{NEURIPS2024_62986e0a,
 author = {Wu, Zhenbang and Dadu, Anant and Nalls, Mike and Faghri, Faraz and Sun, Jimeng},
 booktitle = {Advances in Neural Information Processing Systems},
 doi = {10.52202/079017-1737},
 editor = {A. Globerson and L. Mackey and D. Belgrave and A. Fan and U. Paquet and J. Tomczak and C. Zhang},
 pages = {54772--54786},
 publisher = {Curran Associates, Inc.},
 title = {Instruction Tuning Large Language Models to Understand Electronic Health Records},
 url = {https://proceedings.neurips.cc/paper_files/paper/2024/file/62986e0a78780fe5f17b495aeded5bab-Paper-Datasets_and_Benchmarks_Track.pdf},
 volume = {37},
 year = {2024}
}

@article{Cui2025TIMERTI,
  title={TIMER: temporal instruction modeling and evaluation for longitudinal clinical records},
  author={Cui, Hejie and Unell, Alyssa and Chen, Bowen and Fries, Jason Alan and Alsentzer, Emily and Koyejo, Sanmi and Shah, Nigam H.},
  journal={npj Digital Medicine},
  year={2025},
  volume={8},
  number={1},
  pages={577},
  doi={10.1038/s41746-025-01965-9}
}

@article{Bachmann2024ExploringTC,
  title={Exploring the capabilities of ChatGPT in women’s health: obstetrics and gynaecology},
  author={Bachmann, Magdalena and Duta, Ioana and Mazey, Emily and Cooke, William and Vatish, Manu and Davis Jones, Gabriel},
  journal={npj Women's Health},
  year={2024},
  volume={2},
  pages={26},
  doi={10.1038/s44294-024-00028-w}
}

@article{ugowski2025ComparativeAO,
  title={Comparative analysis of ChatGPT 3.5 and ChatGPT 4 obstetric and gynecological knowledge},
  author={Ługowski, Franciszek and Babińska, Julia and Ludwin, Artur and Stanirowski, Paweł Jan},
  journal={Scientific Reports},
  year={2025},
  volume={15},
  number={1},
  pages={21133},
  doi={10.1038/s41598-025-08424-1}
}

@inproceedings{srikanth-etal-2024-pregnant,
    title = "Pregnant Questions: The Importance of Pragmatic Awareness in Maternal Health Question Answering",
    author = "Srikanth, Neha  and
      Sarkar, Rupak  and
      Mane, Heran  and
      Aparicio, Elizabeth  and
      Nguyen, Quynh  and
      Rudinger, Rachel  and
      Boyd-Graber, Jordan",
    editor = "Duh, Kevin  and
      Gomez, Helena  and
      Bethard, Steven",
    booktitle = "Proceedings of the 2024 Conference of the North American Chapter of the Association for Computational Linguistics: Human Language Technologies (Volume 1: Long Papers)",
    month = jun,
    year = "2024",
    address = "Mexico City, Mexico",
    publisher = "Association for Computational Linguistics",
    url = "https://aclanthology.org/2024.naacl-long.403/",
    doi = "10.18653/v1/2024.naacl-long.403",
    pages = "7253--7268"
}

@misc{Anthropic2025ClaudeHaiku45,
  author       = {{Anthropic}},
  title        = {Introducing Claude Haiku 4.5},
  year         = {2025},
  month        = oct,
  day          = {15},
  howpublished = {\url{https://www.anthropic.com/news/claude-haiku-4-5}},
  note         = {Accessed: 2026-05-17}
}

@misc{OpenAI2026GPT54MiniNano,
  author       = {{OpenAI}},
  title        = {Introducing GPT-5.4 Mini and Nano},
  year         = {2026},
  month        = mar,
  day          = {17},
  howpublished = {\url{https://openai.com/index/introducing-gpt-5-4-mini-and-nano/}},
  note         = {Accessed: 2026-05-17}
}

@misc{Google2025Gemini3Flash,
  author       = {Doshi, Tulsee},
  title        = {Gemini 3 Flash: Frontier Intelligence Built for Speed},
  year         = {2025},
  month        = dec,
  day          = {17},
  howpublished = {\url{https://blog.google/products-and-platforms/products/gemini/gemini-3-flash/}},
  note         = {Google Blog. Accessed: 2026-05-17}
}

@misc{deepseekai2026deepseekv4,
      title={DeepSeek-V4: Towards Highly Efficient Million-Token Context Intelligence},
      author={DeepSeek-AI and Anyi Xu and Bangcai Lin and Bing Xue and Bingxuan Wang and Bingzheng Xu and Bochao Wu and Bowei Zhang and Chaofan Lin and Chen Dong and Chenchen Ling and Chengda Lu and Chenggang Zhao and Chengqi Deng and Chengyu Hou and Chenhao Xu and Chenze Shao and Chong Ruan and Conner Sun and Damai Dai and Daya Guo and Dejian Yang and Deli Chen and Donghao Li and Dongjie Ji and Erhang Li and Fang Wei and Fangyun Lin and Fangzhou Yuan and Feiyu Xia and Fucong Dai and Guangbo Hao and Guanting Chen and Guoai Cao and Guolai Meng and Guowei Li and Han Yu and Han Zhang and Hanwei Xu and Hao Li and Haofen Liang and Haoling Zhang and Haoming Luo and Haoran Wei and Haotian Yuan and Haowei Zhang and Haowen Luo and Haoyu Chen and Haozhe Ji and Hengqing Zhang and Honghui Ding and Hongxuan Tang and Huanqi Cao and Huazuo Gao and Hui Qu and Hui Zeng and J Yang and JQ Zhu and Jia Luo and Jia Song and Jia Yu and Jialiang Huang and Jialu Cai and Jian Liang and Jiangting Zhou and Jiasheng Ye and Jiashi Li and Jiaxin Xu and Jiewen Hu and Jieyu Yang and Jin Chen and Jin Yan and Jingchang Chen and Jingli Zhou and Jingting Xiang and Jingyang Yuan and Jingyuan Cheng and Jingzi Zhou and Jinhua Zhu and Jiping Yu and Joseph Sun and Jun Ran and Junguang Jiang and Junjie Qiu and Junlong Li and Junmin Zheng and Junxiao Song and Kai Dong and Kaige Gao and Kang Guan and Kexing Zhou and Kezhao Huang and Kuai Yu and Lean Wang and Lecong Zhang and Lei Wang and Leyi Xia and Li Zhang and Liang Zhao and Lihua Guo and Lingxiao Luo and Linwang Ma and Linyan Zhu and Litong Wang and Liyu Cai and Liyue Zhang and Longhao Chen and MS Di and MY Xu and Max Mei and Miaojun Wang and Mingchuan Zhang and Minghua Zhang and Minghui Tang and Mingming Li and Mingxu Zhou and Minmin Han and Ning Wang and Panpan Huang and Panpan Wang and Peixin Cong and Peiyi Wang and Peng Zhang and Qiancheng Wang and Qihao Zhu and Qingyang Li and Qinyu Chen and Qiushi Du and Qiwei Jiang and Rui Tian and Ruifan Xu and Ruijie Lu and Ruiling Xu and Ruiqi Ge and Ruisong Zhang and Ruizhe Pan and Runji Wang and Runqian Chen and Runqiu Yin and Runxin Xu and Ruomeng Shen and Ruoyu Zhang and Ruyi Chen and SH Liu and Shanghao Lu and Shangmian Sun and Shangyan Zhou and Shanhuang Chen and Shaofei Cai and Shaoheng Nie and Shaoqing Wu and Shaoyuan Chen and Shengding Hu and Shengyu Liu and Shiqiang Hu and Shirong Ma and Shiyu Wang and Shuiping Yu and Shunfeng Zhou and Shuting Pan and Shuying Yu and Songyang Zhou and Tao Ni and Tao Yun and Tian Jin and Tian Pei and Tian Ye and Tianle Lin and Tianran Ji and Tianyi Cui and Tianyuan Yue and Tingting Yu and Tun Wang and W Zhang and WL Xiao and Wangding Zeng and Wei An and Weilin Zhao and Wen Liu and Wenfeng Liang and Wenjie Pang and Wenjing Luo and Wenjing Yao and Wenjun Gao and Wenkai Yang and Wenlve Huang and Wenqing Hou and Wentao Zhang and Wenting Ma and Xi Gao and Xiang He and Xiangwen Wang and Xianzu Wang and Xiao Bi and Xiaodong Liu and Xiaohan Wang and Xiaokang Chen and Xiaokang Zhang and Xiaotao Nie and Xiaowen Sun and Xiaoxiang Wang and Xin Cheng and Xin Liu and Xin Xie and Xingchao Liu and Xingchen Liu and Xingkai Yu and Xingyou Li and Xinyu Yang and Xinyu Zhang and Xu Chen and Xuanyu Wang and Xuecheng Su and Xueyin Chen and Xuheng Lin and Xuwei Fu and YC Yan and YQ Wang and YW Ma and Yanfeng Luo and Yang Zhang and Yanhong Xu and Yanru Ma and Yanwen Huang and Yao Li and Yao Li and Yao Xu and Yao Zhao and Yaofeng Sun and Yaohui Wang and Yi Qian and Yi Shao and Yi Yu and Yichao Zhang and Yifan Ding and Yifan Shi and Yijia Wu and Yiliang Xiong and Yiling Ma and Ying He and Ying Tang and Ying Zhou and Yingjia Luo and Yinmin Zhong and Yishi Piao and Yisong Wang and Yixiang Zhang and Yixiao Chen and Yixuan Tan and Yixuan Wei and Yiyang Ma and Yiyuan Liu and Yonglun Yang and Yongqiang Guo and Yongtong Wu and Yu Wu and YuKun Li and Yuan Cheng and Yuan Ou and Yuanfan Xu and Yuanhao Li and Yuduan Wang and Yuehan Yang and Yuer Xu and Yuhan Wu and Yuhao Meng and Yuheng Zou and Yukun Zha and Yunfan Xiong and Yupeng Chen and Yuping Lin and Yuqian Cao and Yuqian Wang and Yushun Zhang and Yuting Yan and Yutong Lin and Yuxian Gu and Yuxiang Luo and Yuxiang You and Yuxuan Liu and Yuxuan Zhou and Yuyang Zhou and Yuzhen Huang and ZF Wu and Zehao Wang and Zehua Zhao and Zehui Ren and Zekai Zhang and Zhangli Sha and Zhe Fu and Zhe Ju and Zhean Xu and Zhenda Xie and Zhengyan Zhang and Zheren Gao and Zhewen Hao and Zhibin Gou and Zhicheng Ma and Zhigang Yan and Zhihong Shao and Zhixian Huang and Zhixuan Chen and Zhiyu Wu and Zhizhou Ren and Zhongyu Wu and Zhuoshu Li and Zhuping Zhang and Zian Xu and Zihao Wang and Zihua Qu and Zihui Gu and Zijia Zhu and Zilin Li and Zipeng Zhang and Ziwei Xie and Ziyi Gao and Ziyi Wan and Zizheng Pan and Zongqing Yao},
      year={2026},
      eprint={2606.19348},
      archivePrefix={arXiv},
      primaryClass={cs.CL},
      url={https://arxiv.org/abs/2606.19348}, 
}

@misc{bai2025qwen3vltechnicalreport,
      title={Qwen3-VL Technical Report}, 
      author={Shuai Bai and Yuxuan Cai and Ruizhe Chen and Keqin Chen and Xionghui Chen and Zesen Cheng and Lianghao Deng and Wei Ding and Chang Gao and Chunjiang Ge and Wenbin Ge and Zhifang Guo and Qidong Huang and Jie Huang and Fei Huang and Binyuan Hui and Shutong Jiang and Zhaohai Li and Mingsheng Li and Mei Li and Kaixin Li and Zicheng Lin and Junyang Lin and Xuejing Liu and Jiawei Liu and Chenglong Liu and Yang Liu and Dayiheng Liu and Shixuan Liu and Dunjie Lu and Ruilin Luo and Chenxu Lv and Rui Men and Lingchen Meng and Xuancheng Ren and Xingzhang Ren and Sibo Song and Yuchong Sun and Jun Tang and Jianhong Tu and Jianqiang Wan and Peng Wang and Pengfei Wang and Qiuyue Wang and Yuxuan Wang and Tianbao Xie and Yiheng Xu and Haiyang Xu and Jin Xu and Zhibo Yang and Mingkun Yang and Jianxin Yang and An Yang and Bowen Yu and Fei Zhang and Hang Zhang and Xi Zhang and Bo Zheng and Humen Zhong and Jingren Zhou and Fan Zhou and Jing Zhou and Yuanzhi Zhu and Ke Zhu},
      year={2025},
      eprint={2511.21631},
      archivePrefix={arXiv},
      primaryClass={cs.CV},
      url={https://arxiv.org/abs/2511.21631}, 
}

@misc{qwen3.5,
    title  = {{Qwen3.5}: Towards Native Multimodal Agents},
    author = {{Qwen Team}},
    month  = {February},
    year   = {2026},
    url    = {https://qwen.ai/blog?id=qwen3.5}
}

@misc{xu2025lingshu,
  title={Lingshu: A Generalist Foundation Model for Unified Multimodal Medical Understanding and Reasoning},
  author={{LASA Team}  and Weiwen Xu and Hou Pong Chan and Long Li and Mahani Aljunied and Ruifeng Yuan and Jianyu Wang and Chenghao Xiao and Guizhen Chen and Chaoqun Liu and Zhaodonghui Li and Yu Sun and Junao Shen and Chaojun Wang and Jie Tan and Deli Zhao and Tingyang Xu and Hao Zhang and Yu Rong},
      year={2025},
      eprint={2506.07044},
      archivePrefix={arXiv},
      primaryClass={cs.CL},
      url={https://arxiv.org/abs/2506.07044}, 
}

@misc{GoogleDeepMind2026Gemma4ModelCard,
  author       = {{Google DeepMind}},
  title        = {Gemma 4 Model Card},
  year         = {2026},
  howpublished = {\url{https://ai.google.dev/gemma/docs/core/model_card_4}},
  note         = {Accessed: 2026-05-17}
}

@misc{microsoft2025phi4minitechnicalreportcompact,
      title={Phi-4-Mini Technical Report: Compact yet Powerful Multimodal Language Models via Mixture-of-LoRAs}, 
      author = {Abouelenin, Abdelrahman and Ashfaq, Atabak and Atkinson, Adam and Awadalla, H. and Bach, Nguyen and Bao, Jianmin and Benhaim, A. and Cai, Martin and Chaudhary, Vishrav and Chen, Congcong and Chen, Dongdong and Chen, Dongdong and Chen, Junkun and Chen, Weizhu and Chen, Yen-Chun and Chen, Yi-ling and Dai, Qi and Dai, Xiyang and Fan, Ruchao and Gao, Mei and Gao, Mingcheng and Garg, Amit and Goswami, Abhishek and Hao, Junheng and Hendy, Amr and Hu, Yuxuan and Jin, Xin and Khademi, Mahmoud and Kim, Dongwoo and Kim, Young Jin and Lee, Gina and Li, Jinyu and Li, Yun-Meng and Liang, Chen and Lin, Xihui and Lin, Zeqi and Liu, Meng-Jie and Liu, Yang and Lopez, Gilsinia and Luo, Chong and Madan, Piyush and Mazalov, V. and Mousavi, Ali and Nguyen, Anh and Pan, Jing and Perez-Becker, D. and Platin, Jacob and Portet, Thomas and Qiu, Kai and Ren, Bo and Ren, Liliang and Roy, Sambuddha and Shang, Ning and Shen, Yelong and Singhal, Saksham and Som, Subhojit and Song, Xiaocheng and Sych, Tetyana and Vaddamanu, Praneetha and Wang, Shuohang and Wang, Yiming and Wang, Zhenghao and Wu, Haibin and Xu, Haoran and Xu, Weijian and Yang, Yifan and Yang, Ziyi and Yu, Donghan and Zabir, I. and Zhang, Jianwen and Zhang, L. and Zhang, Yunan and Zhou, Xiren},
      year={2025},
      eprint={2503.01743},
      archivePrefix={arXiv},
      primaryClass={cs.CL},
      url={https://arxiv.org/abs/2503.01743}, 
}

@InProceedings{pmlr-v267-lin25n,
  title =   {{H}ealth{GPT}: A Medical Large Vision-Language Model for Unifying Comprehension and Generation via Heterogeneous Knowledge Adaptation},
  author =       {Lin, Tianwei and Zhang, Wenqiao and Li, Sijing and Yuan, Yuqian and Yu, Binhe and Li, Haoyuan and He, Wanggui and Jiang, Hao and Li, Mengze and Xiaohui, Song and Tang, Siliang and Xiao, Jun and Lin, Hui and Zhuang, Yueting and Ooi, Beng Chin},
  booktitle =   {Proceedings of the 42nd International Conference on Machine Learning},
  pages =   {37975--37995},
  year =   {2025},
  editor =   {Singh, Aarti and Fazel, Maryam and Hsu, Daniel and Lacoste-Julien, Simon and Berkenkamp, Felix and Maharaj, Tegan and Wagstaff, Kiri and Zhu, Jerry},
  volume =   {267},
  series =   {Proceedings of Machine Learning Research},
  month =   {13--19 Jul},
  publisher =    {PMLR},
  url =   {https://proceedings.mlr.press/v267/lin25n.html}
}

@misc{jiang2025hulumedtransparentgeneralistmodel,
      title={Hulu-Med: A Transparent Generalist Model towards Holistic Medical Vision-Language Understanding}, 
      author={Songtao Jiang and Yuan Wang and Sibo Song and Tianxiang Hu and Chenyi Zhou and Bin Pu and Yan Zhang and Zhibo Yang and Yang Feng and Joey Tianyi Zhou and Jin Hao and Zijian Chen and Ruijia Wu and Tao Tang and Junhui Lv and Hongxia Xu and Hongwei Wang and Jun Xiao and Bin Feng and Fudong Zhu and Kenli Li and Weidi Xie and Jimeng Sun and Jian Wu and Zuozhu Liu},
      year={2025},
      eprint={2510.08668},
      archivePrefix={arXiv},
      primaryClass={cs.CV},
      url={https://arxiv.org/abs/2510.08668}, 
}

@misc{sellergren2026medgemma,
  title={MedGemma 1.5 Technical Report},
author={Andrew Sellergren and Chufan Gao and Fereshteh Mahvar and Timo Kohlberger and Fayaz Jamil and Madeleine Traverse and Alberto Tono and Bashir Sadjad and Lin Yang and Charles Lau and Liron Yatziv and Tiffany Chen and Bram Sterling and Kenneth Philbrick and Richa Tiwari and Yun Liu and Madhuram Jajoo and Chandrashekar Sankarapu and Swapnil Vispute and Harshad Purandare and Abhishek Bijay Mishra and Sam Schmidgall and Tao Tu and Anil Palepu and Chunjong Park and Tim Strother and Rahul Thapa and Yong Cheng and Preeti Singh and Kat Black and Yossi Matias and Katherine Chou and Avinatan Hassidim and Kavi Goel and Joelle Barral and Tris Warkentin and Shravya Shetty and Dale Webster and Sunny Virmani and David F. Steiner and Can Kirmizibayrak and Daniel Golden},
      year={2026},
      eprint={2604.05081},
      archivePrefix={arXiv},
      primaryClass={cs.AI},
      url={https://arxiv.org/abs/2604.05081}, 
}

@misc{Anthropic2026ClaudeOpus46,
  author       = {{Anthropic}},
  title        = {Introducing Claude Opus 4.6},
  year         = {2026},
  month        = feb,
  howpublished = {\url{https://www.anthropic.com/news/claude-opus-4-6}},
  note         = {Accessed: 2026-05-25}
}

@misc{OpenAI2026GPT53Codex,
  author       = {{OpenAI}},
  title        = {Introducing GPT-5.3-Codex},
  year         = {2026},
  month        = feb,
  howpublished = {\url{https://openai.com/index/introducing-gpt-5-3-codex/}},
  note         = {Accessed: 2026-05-25}
}

@misc{Anthropic2026ClaudeSonnet46,
  author       = {{Anthropic}},
  title        = {Introducing Claude Sonnet 4.6},
  year         = {2026},
  month        = feb,
  day          = {17},
  howpublished = {\url{https://www.anthropic.com/news/claude-sonnet-4-6}},
  note         = {Accessed: 2026-08-27}
}

@misc{OpenAI2026GPT55,
  author       = {{OpenAI}},
  title        = {Introducing GPT-5.5},
  year         = {2026},
  month        = apr,
  day          = {23},
  howpublished = {\url{https://openai.com/index/introducing-gpt-5-5/}},
  note         = {Accessed: 2026-08-27}
}

@misc{xAI2026Grok45,
  author       = {{xAI}},
  title        = {Introducing Grok 4.5},
  year         = {2026},
  month        = jul,
  day          = {16},
  howpublished = {\url{https://x.ai/news/grok-4-5}},
  note         = {Accessed: 2026-08-27}
}

@misc{qwen3embedding,
  title={Qwen3 Embedding: Advancing Text Embedding and Reranking Through Foundation Models},
      author={Yanzhao Zhang and Mingxin Li and Dingkun Long and Xin Zhang and Huan Lin and Baosong Yang and Pengjun Xie and An Yang and Dayiheng Liu and Junyang Lin and Fei Huang and Jingren Zhou},
      year={2025},
      eprint={2506.05176},
      archivePrefix={arXiv},
      primaryClass={cs.CL},
      url={https://arxiv.org/abs/2506.05176}, 
}

@misc{bao2026medrcubemultidimensionalframeworkfinegrained,
      title={MedRCube: A Multidimensional Framework for Fine-Grained and In-Depth Evaluation of MLLMs in Medical Imaging}, 
      author={Zhijie Bao and Fangke Chen and Licheng Bao and Chenhui Zhang and Wei Chen and Jiajie Peng and Zhongyu Wei},
      year={2026},
      eprint={2604.13756},
      archivePrefix={arXiv},
      primaryClass={cs.CL},
      url={https://arxiv.org/abs/2604.13756}, 
}

\clearpage
\appendix
\section{Data Construction}
\label{app:data-construction}

This appendix presents the complete construction pipeline summarized in Section~\ref{sec:data-construction} and Figure~\ref{fig:construction-pipeline}, together with an independent clinician audit of the benchmark questions and gold answers.
Specifically, Appendix~\ref{app:data-rule-sources} describes the EHR and clinical rule sources, Appendix~\ref{app:construction-pipeline} details the case construction pipeline, and Appendix~\ref{app:clinician-validation} reports the independent clinician audit.

\subsection{Data and Clinical Rule Sources}
\label{app:data-rule-sources}

Our benchmark is built from 976 real pregnancy EHR histories.
For each patient $p$, the longitudinal EHR history $\mathcal{H}_p$ comes from a real obstetric and gynecologic clinical cohort at a single-center tertiary hospital, covering the pregnancy process from early pregnancy registration to pre-delivery assessment.
Each timestamped record $e_i$ contains heterogeneous clinical events, including diagnoses, laboratory tests, vital signs, ultrasound reports, clinical notes, medication records, and screening items.

Based on these EHR histories, we construct clinical decision-point samples $c=(x,q)$.
For a decision point at time $t_k$, the EHR input $x=\mathcal{H}_{p,\leq k}$ consists only of the patient records available up to that point, while all later records are excluded to prevent future information leakage.
The average length of the full pre-decision EHR input $x$ is approximately 30K tokens when tokenized by Qwen3.5, making the benchmark challenging for evidence utilization over long contexts and longitudinal information integration.

Clinical rules $r$ are derived from textbooks and clinical guidelines.
Each rule $r$ defines a type of clinical decision point and provides the basis for constructing the question $q=(m,\mathcal{E},r)$, where $m$ is the multiple-choice question instance and $\mathcal{E} \subseteq x$ is the supporting evidence extracted from the available EHR input.
In this way, the rule source determines the clinical decision to be evaluated, while the patient EHR history determines what evidence is available at the corresponding decision point.

\subsection{Case Construction Pipeline}
\label{app:construction-pipeline}

The construction pipeline consists of two stages: decision-point mining and question construction with grounding.
Starting from textbooks, clinical guidelines, and longitudinal EHR histories, the pipeline first mines specialty-specific clinical rules, matches them to patient EHR histories, and then performs multi-stage review, question drafting, grounding verification, and artifact generation.

Inspired by MIMIC-Instr~\citep{NEURIPS2024_62986e0a}, which constructs EHR-grounded instructions from structured EHR events using programmatic scripts and LLM rewriting, we adapt this paradigm to rule-grounded obstetric and gynecologic clinical decision-making.
Different from general EHR instruction generation, our construction pipeline is centered on clinically meaningful decision points.
It introduces LLM-based rule mining from specialty knowledge sources, programmatic high-recall candidate screening, and multi-stage agent-based gating to ensure that each case is medically valid, temporally grounded, and answerable from pre-decision EHR evidence.

We construct evaluation cases through four steps.

\begin{enumerate}
\item \textbf{Longitudinal EHR history construction.}
We first align raw obstetric and gynecologic EHRs from different modalities and event sources onto a unified gestational-age timeline.
Each patient's records are converted into a patient-centered longitudinal EHR history ordered by time.
This process integrates structured events, such as laboratory tests, vital signs, diagnoses, prescriptions, and screening items, with semi-structured or unstructured records, such as ultrasound reports and clinical notes.

\item \textbf{Clinical rule mining.}
We use GPT-5.3-Codex~\citep{OpenAI2026GPT53Codex} to extract traceable clinical decision rules from obstetrics and gynecology textbooks and clinical guidelines.
Each rule specifies the triggering context, applicable gestational-age or temporal window, required evidence fields, evidence-complexity level, management action, candidate branches, question template, and supporting medical background.
We further use Claude Opus 4.6~\citep{Anthropic2026ClaudeOpus46} to assign each rule to one of three evidence-complexity levels: L1, L2, or L3.

\item \textbf{Programmatic candidate screening with agent-based gating.}
For each clinical rule, we implement a Python screening program to identify patient-rule candidates from the longitudinal EHR history.
This programmatic step serves as a high-recall initial filter.
We then use a GPT-5.3-Codex-based agent~\citep{OpenAI2026GPT53Codex} to review each candidate, check whether the predicted branch is consistent with the patient's pre-decision EHR evidence, and determine whether the case corresponds to a valid clinical decision point.

\item \textbf{Question construction and multi-stage verification.}
For each accepted decision point, an LLM drafts an examination-style clinical question and candidate options based on the matched clinical-rule branch, which determines the gold-answer direction.
We then use GPT-5.4-mini-based verification agents~\citep{OpenAI2026GPT54MiniNano} to check three dimensions: whether the case satisfies the intended clinical rule, whether it forms a valid current-time reasoning question, and whether it is answerable from information available before the decision point.
The grounding agent further verifies the supporting evidence, confirms the pre-decision cutoff, collects evidence targets, and records grounding information for constructing the \textbf{Evidence-only}, \textbf{Visit-level EHR}, and \textbf{History-level EHR} input scopes.
\end{enumerate}

The construction pipeline is designed to produce \textbf{verifiable clinical decision-point cases}, each grounded in a real patient EHR history, a traceable clinical rule, a pre-decision information boundary, and explicit supporting evidence.
This design enables ObGynLongBench to evaluate both rule application under the \textbf{Evidence-only} setting and evidence utilization from full pre-decision EHR histories.
For each instance, the gold-answer direction is fixed by the matched clinical-rule branch and verified against the patient's pre-decision EHR.
The benchmark answers and evaluation scores are therefore objectively defined.
Evaluation uses fixed-choice MCQ accuracy, avoiding the potential circularity and model-family bias associated with open-ended generation or LLM-based subjective evaluation.
The independent clinician audit in Appendix~\ref{app:clinician-validation} provides additional validation of the benchmark questions and gold answers.

\begin{table}[t]
\centering
\footnotesize
\setlength{\tabcolsep}{3pt}
\renewcommand{\arraystretch}{1.05}
\begin{tabularx}{\columnwidth}{@{}Xcc@{}}
\toprule
\textbf{Clinician / statistic} &
\textbf{\shortstack{Gold-answer\\correctness}} &
\textbf{\shortstack{Question\\quality}} \\
\midrule
Clinician 1 & 95/2/3 & 97/2/1 \\
Clinician 2 & 93/2/5 & 97/3/0 \\
Clinician 3 & 98/1/1 & 100/0/0 \\
\textbf{Aggregate result} & \textbf{96/1/3} & \textbf{98/2/0} \\
At least two clinicians agreed & 99\% & 100\% \\
All three clinicians agreed & 90\% & 96\% \\
Gwet's AC1 & 0.927 & 0.973 \\
\bottomrule
\end{tabularx}
\caption{Results of the independent clinician audit on 100 randomly sampled benchmark instances. For gold-answer correctness, values are reported in the order Correct/Incorrect/Uncertain. For question clarity and clinical meaningfulness, values are reported in the order Clear and meaningful/Ambiguous/Unreasonable. Clinician-level and aggregate results are counts out of 100.}
\label{tab:clinician-audit}
\end{table}

\begin{table*}[t]
\centering
\footnotesize
\resizebox{0.9\textwidth}{!}{
\begin{tabular}{lc|c|c|c}
\hline
\rowcolor[HTML]{D9E2F3}
\multicolumn{1}{c|}{\cellcolor[HTML]{D9E2F3}\textbf{Model}} & \multicolumn{1}{c|}{\cellcolor[HTML]{D9E2F3}\textbf{Context window}} & \textbf{Evidence-only Overall} & \textbf{History-level EHR Overall} & \textbf{Visit-level EHR Overall} \\ \hline
\rowcolor[HTML]{F2F2F2}
\multicolumn{5}{c}{\cellcolor[HTML]{F2F2F2}\textbf{Commercial LLMs (Flash Level)}} \\ \hline
Claude-Haiku-4.5 & 256k & \accpm{75.5}{2.2} & \accpm{59.8}{2.6} & \accpm{58.5}{2.5} \\
GPT-5.4-mini & 400k & \accpm{77.4}{2.2} & \accpm{64.9}{2.4} & \accpm{62.3}{2.4} \\
Gemini-3-Flash & 1000k & \accpm{79.1}{2.1} & \accpm{68.7}{2.4} & \accpm{67.1}{2.4} \\
DeepSeek-V4-Flash & 1000k & \accpm{78.8}{2.1} & \accpm{64.4}{2.5} & \accpm{61.2}{2.5} \\ \hline
\rowcolor[HTML]{F2F2F2}
\multicolumn{5}{c}{\cellcolor[HTML]{F2F2F2}\textbf{Long-context open-source LLMs}} \\ \hline
Qwen3-VL-4B & 262k & \accpm{72.1}{2.3} & \accpm{59.8}{2.5} & \accpm{54.7}{2.6} \\
Qwen3-VL-8B & 262k & \accpm{74.3}{2.3} & \accpm{61.8}{2.5} & \accpm{58.4}{2.5} \\
Qwen3.5-4B & 262k & \accpm{75.4}{2.2} & \accpm{59.4}{2.6} & \accpm{52.7}{2.6} \\
Qwen3.5-9B & 262k & \accpm{73.1}{2.2} & \accpm{59.3}{2.6} & \accpm{55.4}{2.5} \\
Qwen3.5-35B-A3B & 262k & \accpm{76.9}{2.3} & \accpm{60.1}{2.5} & \accpm{55.7}{2.5} \\
Gemma-4-E4B & 131k & \accpm{68.1}{2.4} & \accpm{52.0}{2.6} & \accpm{47.4}{2.5} \\
Gemma-4-26B-A4B & 262k & \accpm{78.2}{2.2} & \accpm{60.7}{2.5} & \accpm{60.6}{2.5} \\
Phi-4-mini & 131k & \accpm{58.7}{2.6} & \accpm{51.1}{2.5} & \accpm{48.6}{2.5} \\ \hline
\rowcolor[HTML]{F2F2F2}
\multicolumn{5}{c}{\cellcolor[HTML]{F2F2F2}\textbf{Long-context Medical LLMs}} \\ \hline
HealthGPT-Pro-4B & 262k & \accpm{69.6}{2.3} & \accpm{57.1}{2.5} & \accpm{54.1}{2.6} \\
HealthGPT-Pro-8B & 262k & \accpm{73.5}{2.3} & \accpm{61.3}{2.6} & \accpm{58.1}{2.5} \\
Hulu-Med-30A3 & 75k & \accpm{74.9}{2.2} & \accpm{53.9}{2.6} & \accpm{55.8}{2.6} \\
Lingshu-7B & 128k & \accpm{72.1}{2.4} & \accpm{59.6}{2.7} & \accpm{60.4}{2.5} \\
MedGemma-1.5-4B & 131k & \accpm{64.0}{2.4} & \accpm{41.3}{2.6} & \accpm{48.1}{2.5} \\
\end{tabular}
}
\caption{Patient-clustered bootstrap uncertainty for overall accuracy under the three input-scope settings. Each entry is reported as $\hat{A}\pm m$ in percentage points, where $m$ is the half-width of a 95\% patient-clustered percentile bootstrap confidence interval (Eqs.~\eqref{eq:acc-boot}--\eqref{eq:acc-margin}).}
\label{tab:bootstrap-margins}
\end{table*}

\subsection{Clinician Validation}
\label{app:clinician-validation}

To assess the reliability of the benchmark, we conducted an independent clinician audit of 100 instances randomly sampled from the 1,500-instance benchmark.
Three obstetrics and gynecology resident physicians with clinical experience participated in the audit, none of whom was involved in benchmark construction.
Each clinician independently reviewed the same 100 instances against the complete EHR history available up to the corresponding pre-decision cutoff.

For each instance, the clinicians were provided with the question and answer options, the designated gold answer, and the complete pre-decision EHR history.
They independently assessed two dimensions using predefined three-category rubrics: (1) gold-answer correctness, categorized as correct, incorrect, or uncertain; and (2) question clarity and clinical meaningfulness, categorized as clear and meaningful, ambiguous and requiring revision, or unreasonable.

For gold-answer correctness, majority voting was applied to the 99 instances for which at least two clinicians agreed.
The single instance with three-way disagreement was conservatively categorized as uncertain.
Overall, 96 gold answers were classified as correct, one as incorrect, and three as uncertain, corresponding to an observed clearly incorrect rate of 1\% in the audit sample.
Based on this clinician audit, we estimate the benchmark-wide rate of clearly incorrect gold answers to be below 5\%.
For question quality, 98 instances were classified as clear and clinically meaningful, two as ambiguous and requiring revision, and none as unreasonable.

Inter-rater agreement was high for both dimensions.
For gold-answer correctness, at least two clinicians agreed on 99\% of the instances, all three agreed on 90\%, and Gwet's AC1 was 0.927.
For question clarity and clinical meaningfulness, the corresponding results were 100\%, 96\%, and 0.973.
We report Gwet's AC1 because the annotations were highly concentrated in one category, under which conventional kappa statistics can be affected by the prevalence paradox.
Overall, the audit provides independent clinician-based evidence supporting the reliability of the benchmark questions and gold answers.

\section{Difficulty Stratification and Metrics}
\label{sec:difficulty-metrics}

To analyze LLM performance under different levels of evidence complexity, we assign clinical rules to three mutually exclusive levels. Cases are then grouped by the rule used to construct them.

\textbf{L1: Single-evidence recognition.}
L1 rules depend on one explicit and locally visible key evidence item, such as a single abnormal indicator, one positive screening result, or a clear diagnosis.

\textbf{L2: Multi-source integration.}
L2 rules require integrating multiple evidence items or modalities within the same visit or local context, such as jointly interpreting laboratory results and ultrasound findings, or using multiple indicators together to support a management choice.

\textbf{L3: Long-horizon reasoning.}
L3 rules require previous visit records, cross-time trends, or historical evidence far from the decision point, such as events more than 7 days before the decision point.
These rules therefore require LLMs to perform temporal anchoring and longitudinal integration over long-range EHR histories.

The stratification is based on \textbf{evidence complexity}, rather than strict medical difficulty, in order to better characterize LLMs' ability to identify and integrate historical evidence in long-context EHRs.

For metrics, we mainly report overall accuracy, and further report stratified accuracy by evidence complexity, input scope, and EHR access strategy.

\section{Statistical Uncertainty and Additional Analyses}
\label{app:fine-grained-analysis}

\subsection{Patient-Clustered Bootstrap Uncertainty for Main Results}
\label{app:bootstrap-uncertainty}

To quantify the statistical uncertainty of model evaluation, we compute patient-clustered percentile bootstrap confidence intervals for each model under each input-scope setting.

For a fixed model and input scope, let $\mathcal{D}_p=\{(y_{pj},\hat{y}_{pj})\}_{j=1}^{n_p}$ denote the decision-point cases contributed by patient $p$, where $y_{pj}$ is the gold answer and $\hat{y}_{pj}$ is the model prediction.
Across $P$ patients, the total number of cases is $N=\sum_{p=1}^{P}n_p$.
The point accuracy is
\begin{equation}
\label{eq:acc-point}
\hat{A}
=
\frac{1}{N}
\sum_{p=1}^{P}
\sum_{j=1}^{n_p}
\mathbb{I}(\hat{y}_{pj}=y_{pj}),
\end{equation}
which is consistent with Eq.~\eqref{eq:acc}.

For each bootstrap replicate $b=1,\ldots,B$, we sample $P$ patients with replacement.
Let $i_1^{(b)},\ldots,i_P^{(b)}$ denote the sampled patient indices and let $N^{(b)}=\sum_{k=1}^{P}n_{i_k^{(b)}}$.
All decision-point cases belonging to each sampled patient are retained.
The bootstrap accuracy is
\begin{equation}
\label{eq:acc-boot}
\hat{A}^{(b)}
=
\frac{1}{N^{(b)}}
\sum_{k=1}^{P}
\sum_{j=1}^{n_{i_k^{(b)}}}
\mathbb{I}\left(\hat{y}_{i_k^{(b)}j}=y_{i_k^{(b)}j}\right).
\end{equation}

We report the two-sided $100(1-\alpha)\%$ percentile confidence interval
\begin{equation}
\label{eq:acc-ci}
\left[Q_{\alpha/2}(\hat{A}^{*}),\;Q_{1-\alpha/2}(\hat{A}^{*})\right],
\end{equation}
where $\hat{A}^{*}$ denotes the empirical distribution of the bootstrap replicates and $Q_q(\hat{A}^{*})$ denotes its $q$-th quantile.
In Table~\ref{tab:bootstrap-margins}, we report the half-width
\begin{equation}
\label{eq:acc-margin}
m
=
\frac{Q_{1-\alpha/2}(\hat{A}^{*})-Q_{\alpha/2}(\hat{A}^{*})}{2},
\end{equation}
so that each result is presented as $\hat{A}\pm m$ in percentage points.
We use $B=10{,}000$ and $\alpha=0.05$, with a fixed random seed for each model--scope pair.

\subsection{Expanded Patient-Level Analysis}
\label{app:expanded-patient-analysis}

We further evaluate model performance across longitudinal decision points at the patient level using all 1,500 cases and all 17 models under the \textbf{History-level EHR} setting.
We define each patient's first decision point as DP1, yielding 976 cases, and group all subsequent decision points as DP2+, yielding 524 cases.
For each model, we test the one-sided hypothesis that DP2+ accuracy is lower than DP1 accuracy.

\begin{table}[htbp]
  \centering
  \footnotesize
  \setlength{\tabcolsep}{2pt}
  \resizebox{\columnwidth}{!}{%
  \begin{tabular}{lrrrr}
    \toprule
    \textbf{Model} & \textbf{DP1 Acc.} & \textbf{DP2+ Acc.} & \textbf{Diff. (pp)} & \textbf{$p$} \\
    \midrule
    Claude-Haiku-4.5    & 61.99 & 55.73 & $-6.26$  & $0.0095^{**}$ \\
    GPT-5.4-mini        & 65.57 & 63.55 & $-2.02$  & $0.2177$ \\
    Gemini-3-Flash      & 69.67 & 66.98 & $-2.69$  & $0.1438$ \\
    DeepSeek-V4-Flash   & 66.29 & 60.88 & $-5.41$  & $0.0192^{**}$ \\
    Qwen3-VL-4B         & 60.45 & 58.59 & $-1.86$  & $0.2419$ \\
    Qwen3-VL-8B         & 63.83 & 58.02 & $-5.82$  & $0.0140^{**}$ \\
    Qwen3.5-4B          & 60.66 & 57.06 & $-3.59$  & $0.0890^{*}$ \\
    Qwen3.5-9B          & 61.17 & 55.73 & $-5.44$  & $0.0208^{**}$ \\
    Qwen3.5-35B-A3B     & 63.93 & 53.05 & $-10.88$ & $<0.0001^{***}$ \\
    Gemma-4-E4B         & 54.30 & 47.71 & $-6.59$  & $0.0073^{**}$ \\
    Gemma-4-26B-A4B     & 63.93 & 54.58 & $-9.35$  & $0.0002^{***}$ \\
    Phi-4-mini          & 54.00 & 45.80 & $-8.19$  & $0.0012^{**}$ \\
    HealthGPT-Pro-4B    & 57.89 & 55.73 & $-2.16$  & $0.2101$ \\
    HealthGPT-Pro-8B    & 63.11 & 57.82 & $-5.29$  & $0.0231^{**}$ \\
    Hulu-Med-30A3       & 56.05 & 49.81 & $-6.24$  & $0.0105^{**}$ \\
    Lingshu-7B          & 60.66 & 57.63 & $-3.02$  & $0.1284$ \\
    MedGemma-1.5-4B     & 43.85 & 36.64 & $-7.21$  & $0.0031^{**}$ \\
    \bottomrule
  \end{tabular}%
  }
  \caption{Expanded patient-level analysis under the \textbf{History-level EHR} setting. Accuracy is reported in percent, and the difference is computed as DP2+ accuracy minus DP1 accuracy. $^{*}p<0.1$, $^{**}p<0.05$, and $^{***}p<0.001$.}
  \label{tab:expanded-patient-analysis}
\end{table}

DP2+ accuracy is lower than DP1 accuracy for all 17 models, with decreases ranging from 1.86 to 10.88 percentage points and an average decrease of 5.41 points.
The decrease is statistically significant at $p<0.1$ for 12 of the 17 models.
These results show that later decision points are generally more challenging across models.
Together with the conditional analysis in Section~\ref{sec:patient-stability}, they further show that model errors across longitudinal decision points are associated within patients.

\subsection{Paired-Bootstrap Uncertainty for EHR Access Strategies}
\label{app:access-bootstrap}

To quantify the statistical uncertainty in the comparison of EHR access strategies, we perform a paired-bootstrap analysis by pairing each strategy with \textbf{Direct} access on the same benchmark cases.
Table~\ref{tab:access-bootstrap} reports the overall accuracy difference relative to \textbf{Direct} access and the half-width of its 95\% paired-bootstrap confidence interval.

\begin{table}[htbp]
  \centering
  \footnotesize
  \setlength{\tabcolsep}{4pt}
  \begin{tabular}{lr}
    \toprule
    \textbf{Access strategy} & \textbf{Accuracy gap vs. Direct (pp)} \\
    \midrule
    Image             & $-6.91 \pm 1.59$ \\
    Static RAG        & $-0.71 \pm 1.44$ \\
    Static RAG + Rec. & $-0.44 \pm 1.47$ \\
    Rolling Summary   & $-1.67 \pm 1.51$ \\
    Agent             & $+4.53 \pm 1.60$ \\
    \bottomrule
  \end{tabular}
  \caption{Paired-bootstrap uncertainty for the EHR access-strategy comparison. Each entry reports the overall accuracy difference relative to \textbf{Direct} access and the half-width of its 95\% paired-bootstrap confidence interval.}
  \label{tab:access-bootstrap}
\end{table}

\subsection{Chapter-level Analysis}
\label{app:chapter-analysis}

This subsection provides a chapter-level analysis under the \textbf{History-level EHR} setting.
We group benchmark cases by obstetric and gynecologic chapters to examine how LLM performance varies across clinical topics when LLMs must answer from the full pre-decision EHR history.

\begin{figure}[t]
  \centering
  \includegraphics[width=\columnwidth]{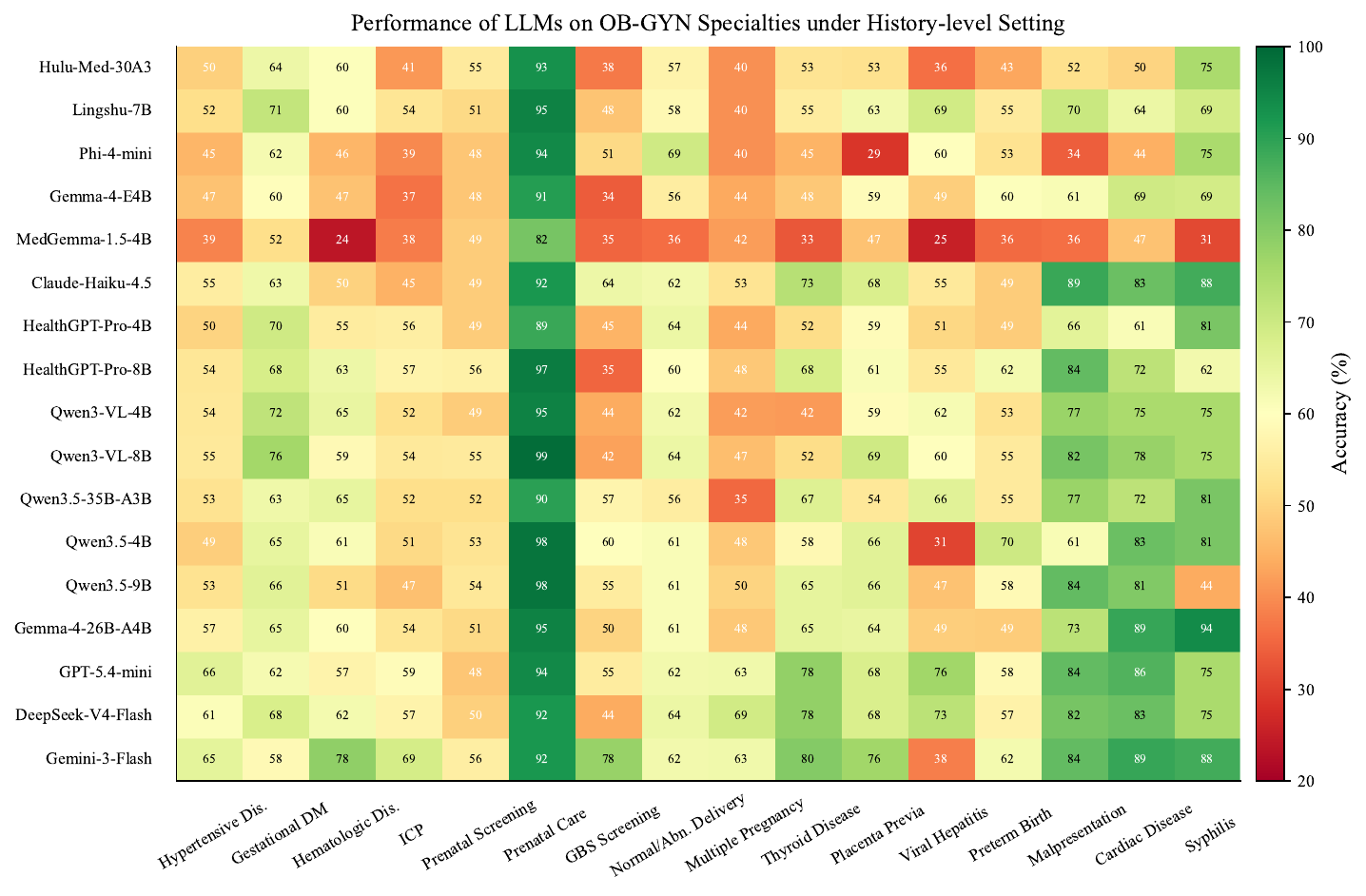}
  \caption{Obstetric and gynecologic chapter-level accuracy heatmap under the \textbf{History-level EHR} setting.}
  \label{fig:chapter}
\end{figure}

Figure~\ref{fig:chapter} shows that LLM performance varies substantially across chapters under \textbf{History-level EHR}.
In the \textbf{Prenatal Care} chapter, LLMs maintain relatively high accuracy.
Clinically, prenatal care usually starts with early pregnancy registration and gestational-age confirmation; many prenatal care tasks have clear gestational-age anchors, repeated visit structures, and relatively standardized examination pathways~\citep{Peahl2020TheEO}.
These features make it easier for LLMs to locate decision-relevant evidence from the pre-decision EHR history.

In contrast, chapters such as \textbf{ICP}, \textbf{Prenatal Screening}, \textbf{GBS Screening}, \textbf{Hypertensive Disorders}, and \textbf{Multiple Pregnancy} show lower accuracy under \textbf{History-level EHR}.
Taking \textbf{ICP} as an example, clinical judgment requires jointly considering pruritus symptoms, total bile acids, liver enzymes, gestational age, and timing of delivery.
SMFM recommends measuring total bile acids and transaminases when ICP is suspected, and determining monitoring and delivery strategies based on bile acid levels and gestational age~\citep{Lee2020SocietyFM}.
Thus, LLMs must not only identify an abnormal laboratory result, but also determine whether it occurs at the relevant gestational age, whether it matches the symptoms, whether it reaches the management threshold, and how it relates to delivery timing.

Similarly, \textbf{GBS Screening} depends on the screening window of 36+0--37+6 weeks of gestation, culture results, and history of GBS bacteriuria or previous neonatal infection~\citep{2020PreventionOG};
\textbf{Hypertensive Disorders} requires distinguishing blood pressure before and after 20 weeks, proteinuria, platelet count, liver and kidney function, and severe features~\citep{2020GestationalHA};
and \textbf{Prenatal Screening} requires integrating gestational age, screening method, fetal number, previous screening results, and subsequent diagnostic pathways~\citep{Rose2020ScreeningFF}.
The correct answers in these chapters are not determined by a single record, but depend jointly on gestational age, repeated examinations, and historical results.
This explains why these chapters place greater demands on evidence utilization, temporal anchoring, and longitudinal integration under the \textbf{History-level EHR} setting.

\section{LLM-as-Judge Robustness}
\label{app:judge-robustness}

DeepSeek-V4-Flash was used as the primary judge for the evidence-utilization scores reported in Figure~\ref{fig:evidence-accuracy-input}(b), following the 0--5 rubric described in Appendix~\ref{app:prompts-judge}.
To assess robustness to judge-model choice, we used three additional judges, Claude-Sonnet-4.6, GPT-5.5, and Grok-4.5, to re-score the same 100 randomly sampled instances using the identical prompt, inputs, and rubric.
Table~\ref{tab:cross-judge-qwk} reports the pairwise quadratic weighted kappa results.

\begin{table}[t]
  \centering
  \footnotesize
  \setlength{\tabcolsep}{3pt}
  \renewcommand{\arraystretch}{1.05}
  \begin{tabular}{lcccc}
    \toprule
    \textbf{Judge} & \textbf{DS-V4} & \textbf{CS-4.6} & \textbf{GPT-5.5} & \textbf{Grok-4.5} \\
    \midrule
    DS-V4    & 1.00 & 0.77 & 0.74 & 0.73 \\
    CS-4.6   & 0.77 & 1.00 & 0.85 & 0.83 \\
    GPT-5.5  & 0.74 & 0.85 & 1.00 & 0.85 \\
    Grok-4.5 & 0.73 & 0.83 & 0.85 & 1.00 \\
    \bottomrule
  \end{tabular}
  \caption{Pairwise quadratic weighted kappa among four LLM judges on the same 100 randomly sampled instances. The judges are DeepSeek-V4-Flash (DS-V4)~\citep{deepseekai2026deepseekv4}, Claude-Sonnet-4.6 (CS-4.6)~\citep{Anthropic2026ClaudeSonnet46}, GPT-5.5~\citep{OpenAI2026GPT55}, and Grok-4.5~\citep{xAI2026Grok45}. All judges used the identical prompt, inputs, and 0--5 evidence-utilization rubric; DS-V4 was the primary judge used in Figure~\ref{fig:evidence-accuracy-input}(b).}
  \label{tab:cross-judge-qwk}
\end{table}

Pairwise QWK ranged from 0.73 to 0.85, with a mean of 0.80 across the four judge models.
These results indicate that the four models produced consistently aligned evidence-utilization scores when applying the same 0--5 rubric.

\section{Computational Budget}
\label{app:computational-budget}

Commercial LLMs are evaluated through their official APIs, so their parameter counts and serving infrastructure are not publicly available.
For open-source and medical LLMs, model sizes are indicated by their released model names or official model cards when available.
Local inference experiments are run with vLLM on NVIDIA H100 80GB GPUs.
Because our experiments combine API-based commercial LLMs and locally served open-source or medical LLMs with different inference infrastructures, a single unified GPU-hour estimate is not directly comparable across model groups.
For EHR access strategies, we report average token cost in Table~\ref{tab:qwen-modes}.

\section{Implementation Details}
\label{app:implementation-details}

We use the Qwen3.5 tokenizer to measure EHR input length.
Local inference for open-source and medical LLMs is performed with vLLM 0.19.1 on NVIDIA H100 80GB GPUs.
For these locally deployed models, we disable thinking mode when applicable and use the generation parameters recommended in each model's official generation configuration.
Commercial LLMs are evaluated through their official APIs, using the default automatic reasoning or thinking configuration when provided by the API.

For \textbf{Rolling Summary}, EHR histories are processed in consecutive 7-day windows.
All EHR access strategies are evaluated on the same benchmark instances using the same three Qwen3.5-series base models.
Because \textbf{Agent} and \textbf{Rolling Summary} involve multiple inference rounds, we do not impose an identical total token budget across strategies.
Instead, Table~\ref{tab:qwen-modes} reports the actual average token cost of each strategy.
All prompt templates are provided in Appendix~\ref{app:prompts}.

\subsection{RAG Implementation Details}
\label{app:rag-details}

Retrieval uses Qwen3-Embedding-4B~\citep{qwen3embedding} embeddings and cosine similarity.
Each pre-decision EHR is split at natural calendar-date boundaries, and chunks do not span multiple dates.
Within each date, records are organized by event type.
The maximum chunk length is 10,000 characters; longer records are split at event-type or event-batch boundaries.

\textbf{Static RAG} uses the question text as the retrieval query and returns the top-10 chunks.
\textbf{Static RAG + Recency} uses the same retrieval setting and applies linear temporal weighting:
\begin{equation}
\label{eq:recency-weighting}
s_i^{\mathrm{final}}
=
s_i^{\mathrm{retrieval}}
\left(
1 + 0.3\frac{i}{n-1}
\right),
\end{equation}
where the retrieved chunks are ordered chronologically, with $i=0,\ldots,n-1$.
The oldest chunk therefore receives a weight of 1.0, whereas the most recent chunk receives a weight of 1.3.

\subsection{Agent Implementation Details}
\label{app:agent-details}

Each call to \texttt{search\_ehr} returns five chunks by default, while the model may request up to ten.
The \textbf{Agent} may issue multiple searches using different queries and runs for at most 20 interaction turns.
It stops when \texttt{submit\_answer} is invoked.
If the maximum number of turns is reached, the model is instructed to submit its final answer immediately.
The Agent prompt is provided in Appendix~\ref{app:prompts-agent}.

\section*{Licenses}

We use publicly available models, benchmarks, and related resources in accordance with their respective licenses or terms of use.
Our evaluation code is publicly available under the MIT License.

\section{Prompts}
\label{app:prompts}

This appendix lists the data-construction, MCQ inference, Agent, and evaluation prompt templates used in ObGynLongBench.
Unless otherwise noted, all single-pass settings share one two-message chat template: one \texttt{system} message and one \texttt{user} message.
The \textbf{Agent} access strategy uses a multi-turn tool-calling loop instead (Appendix~\ref{app:prompts-agent}).
The \textbf{Image} access strategy sends the pre-decision EHR as rendered page images to a vision-language model (Appendix~\ref{prompt:image}).
Each case is first converted to a multiple-choice question string (\texttt{format\_question}): the question stem followed by options A--D on separate lines.
The original deployed construction and MCQ prompts are written in Chinese; we provide English translations below.
Model rationales are also requested in Chinese in the MCQ output JSON.
All prompt figures are collected at the end of this appendix.

\newtcolorbox{promptbox}[1]{
  enhanced,
  colback=white,
  colframe=gray!62,
  boxrule=0.45pt,
  arc=0pt,
  title={#1},
  coltitle=black,
  colbacktitle=gray!17,
  fonttitle=\bfseries\footnotesize,
  left=6pt,
  right=6pt,
  top=3pt,
  bottom=3pt,
  before skip=4pt,
  after skip=2pt,
  fontupper=\scriptsize,
  width=\textwidth
}

\subsection{Data Construction Prompts}
\label{app:prompts-construction}

Figure~\ref{fig:construction-pipeline} summarizes the overall construction pipeline (Appendix~\ref{app:construction-pipeline}).
This subsection lists the LLM prompts used in rule mining, evidence-complexity assignment, candidate branch classification, and question construction with verification.
Candidate branch classification and grounding verification use multi-turn agents; we show only the core \texttt{system} instructions in the figures below.

\begin{figure}[H]
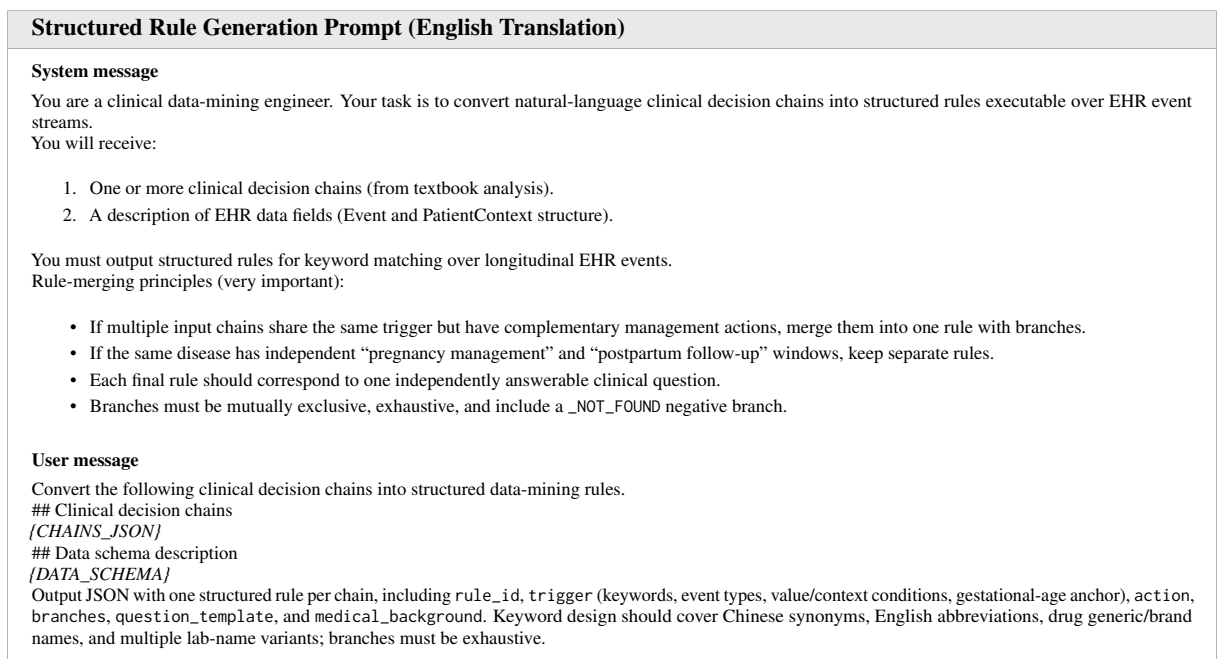

  \centering
  \begin{promptbox}{Structured Rule Generation Prompt (English Translation)}
    \textbf{System message}\par\smallskip
    You are a clinical data-mining engineer. Your task is to convert natural-language clinical decision chains into structured rules executable over EHR event streams.

    You will receive:
    \begin{enumerate}\setlength{\itemsep}{2pt}\setlength{\parskip}{0pt}
      \item One or more clinical decision chains (from textbook analysis).
      \item A description of EHR data fields (Event and PatientContext structure).
    \end{enumerate}

    You must output structured rules for keyword matching over longitudinal EHR events.

    Rule-merging principles (very important):
    \begin{itemize}\setlength{\itemsep}{2pt}\setlength{\parskip}{0pt}
      \item If multiple input chains share the same trigger but have complementary management actions, merge them into one rule with branches.
      \item If the same disease has independent ``pregnancy management'' and ``postpartum follow-up'' windows, keep separate rules.
      \item Each final rule should correspond to one independently answerable clinical question.
      \item Branches must be mutually exclusive, exhaustive, and include a \texttt{\_NOT\_FOUND} negative branch.
    \end{itemize}

    \smallskip
    \textbf{User message}\par\smallskip
    Convert the following clinical decision chains into structured data-mining rules.

    \#\# Clinical decision chains\\
    \textit{\{CHAINS\_JSON\}}

    \#\# Data schema description\\
    \textit{\{DATA\_SCHEMA\}}

    Output JSON with one structured rule per chain, including \texttt{rule\_id}, \texttt{trigger} (keywords, event types, value/context conditions, gestational-age anchor), \texttt{action}, \texttt{branches}, \texttt{question\_template}, and \texttt{medical\_background}.
    Keyword design should cover Chinese synonyms, English abbreviations, drug generic/brand names, and multiple lab-name variants; branches must be exhaustive.
  \end{promptbox}
  \caption{Prompt for converting discovered decision chains into structured executable rules.}
  \label{fig:prompt-rule-definition}
\end{figure}

\paragraph{Clinical rule mining.}
Rule mining proceeds in two LLM steps: (1)~discover clinical decision chains from textbook/guideline text (Figure~\ref{fig:prompt-chain-discovery}), and (2)~convert each chain into a structured executable rule (Figure~\ref{fig:prompt-rule-definition}).

\begin{figure}[H]
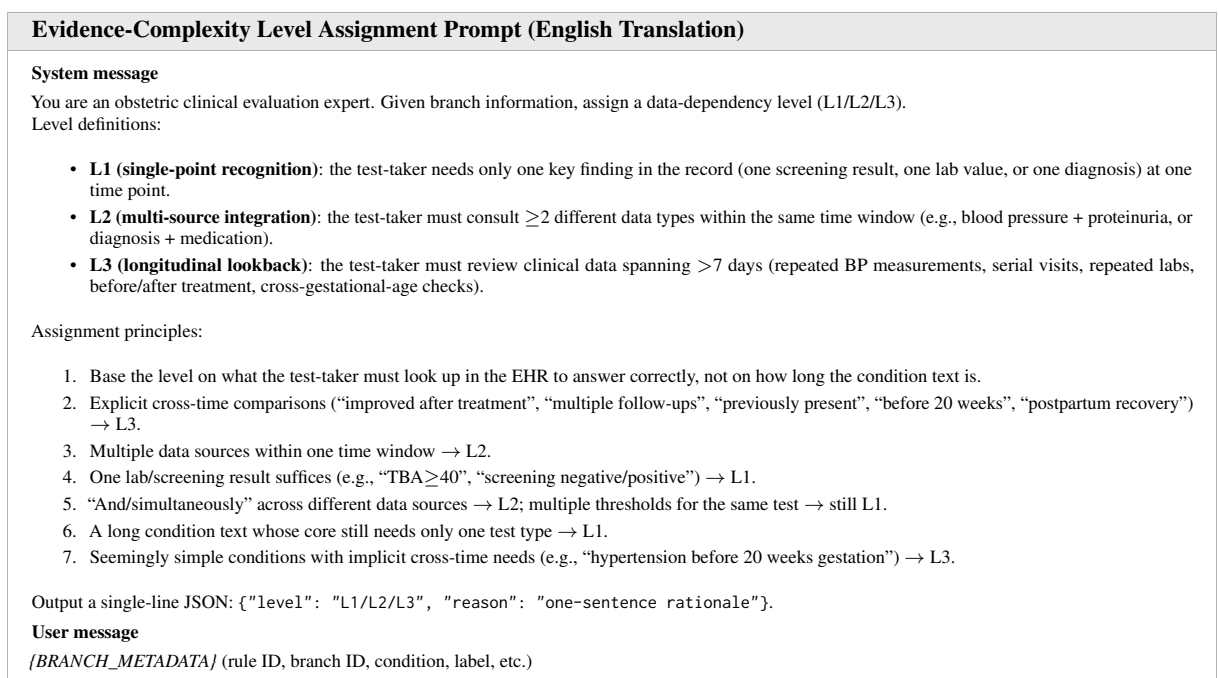

  \centering
  \begin{promptbox}{Evidence-Complexity Level Assignment Prompt (English Translation)}
    \textbf{System message}\par\smallskip
    You are an obstetric clinical evaluation expert. Given branch information, assign a data-dependency level (L1/L2/L3).

    Level definitions:
    \begin{itemize}\setlength{\itemsep}{2pt}\setlength{\parskip}{0pt}
      \item \textbf{L1 (single-point recognition)}: the test-taker needs only one key finding in the record (one screening result, one lab value, or one diagnosis) at one time point.
      \item \textbf{L2 (multi-source integration)}: the test-taker must consult $\geq$2 different data types within the same time window (e.g., blood pressure + proteinuria, or diagnosis + medication).
      \item \textbf{L3 (longitudinal lookback)}: the test-taker must review clinical data spanning $>$7 days (repeated BP measurements, serial visits, repeated labs, before/after treatment, cross-gestational-age checks).
    \end{itemize}

    Assignment principles:
    \begin{enumerate}\setlength{\itemsep}{2pt}\setlength{\parskip}{0pt}
      \item Base the level on what the test-taker must look up in the EHR to answer correctly, not on how long the condition text is.
      \item Explicit cross-time comparisons (``improved after treatment'', ``multiple follow-ups'', ``previously present'', ``before 20 weeks'', ``postpartum recovery'') $\rightarrow$ L3.
      \item Multiple data sources within one time window $\rightarrow$ L2.
      \item One lab/screening result suffices (e.g., ``TBA$\geq$40'', ``screening negative/positive'') $\rightarrow$ L1.
      \item ``And/simultaneously'' across different data sources $\rightarrow$ L2; multiple thresholds for the same test $\rightarrow$ still L1.
      \item A long condition text whose core still needs only one test type $\rightarrow$ L1.
      \item Seemingly simple conditions with implicit cross-time needs (e.g., ``hypertension before 20 weeks gestation'') $\rightarrow$ L3.
    \end{enumerate}

    Output a single-line JSON: \texttt{\{"level": "L1/L2/L3", "reason": "one-sentence rationale"\}}.

    \smallskip
    \textbf{User message}\par\smallskip
    \textit{\{BRANCH\_METADATA\}} (rule ID, branch ID, condition, label, etc.)
  \end{promptbox}
  \caption{Prompt for assigning evidence-complexity levels (L1/L2/L3) to rule branches.}
  \label{fig:prompt-level-assignment}
\end{figure}

\paragraph{Evidence-complexity level assignment.}
Each rule branch is assigned to L1, L2, or L3 based on the evidence a test-taker must locate in the EHR to answer correctly (Figure~\ref{fig:prompt-level-assignment}).

\paragraph{Candidate branch classification.}
After programmatic high-recall screening, an agent reviews each patient--rule candidate and assigns a branch ID consistent with pre-decision EHR evidence (Figure~\ref{fig:prompt-candidate-classification}).

\paragraph{Question construction and verification.}
Accepted candidates pass four LLM steps: chain audit (Figure~\ref{fig:prompt-chain-audit}), draft question generation (Figure~\ref{fig:prompt-draft-question}), grounding verification (Figure~\ref{fig:prompt-grounding}), and final rewrite (Figure~\ref{fig:prompt-final-rewrite}).
These steps are implemented as separate prompts rather than three independent verification agents.

\subsection{Inference and Evaluation Prompts}
\label{app:prompts-inference-eval}

The prompts below are used at benchmark inference and analysis time.

\subsection{Shared MCQ Inference Template}
\label{app:prompts-inference}

Figure~\ref{fig:prompt-mcq-shared} shows the shared template used by \textbf{Evidence-only}, \textbf{Visit-level EHR}, and \textbf{History-level EHR}.
Table~\ref{tab:prompt-variations} summarizes the setting-specific \textit{\{SCOPE\_INSTRUCTION\}} and user input layout; \textbf{Static RAG}, \textbf{Static RAG + Recency}, and \textbf{Image} reuse the same system-message template with different scope instructions.
\textbf{Image} renders the full pre-decision EHR as page-level images and uses a vision-language input format (Figure~\ref{fig:prompt-image}).
\textbf{Rolling Summary} uses a separate rolling-summarization prompt for intermediate windows and a final MCQ prompt on the last window (Figure~\ref{fig:prompt-rolling-summarize} and Figure~\ref{fig:prompt-rolling-final}).
\textbf{Agent} uses a multi-turn tool-calling prompt under the \textbf{History-level EHR} setting (Figure~\ref{fig:prompt-agent}).

\subsection{Static RAG User Message}
\label{prompt:static-rag}

\textbf{Static RAG} and \textbf{Static RAG + Recency} reuse the system-message template with \textit{\{SCOPE\_INSTRUCTION\}} from Table~\ref{tab:prompt-variations}, but use the user-message layout in Figure~\ref{fig:prompt-static-rag}.

\subsection{Image Prompt}
\label{prompt:image}

Under \textbf{Image}, the full pre-decision EHR text is paginated and rendered as grayscale PNG page images.
The vision-language model reuses the same system-message template with \textit{\{SCOPE\_INSTRUCTION\}} from Table~\ref{tab:prompt-variations}, but receives rendered page images together with the formatted question in a multi-modal user message (Figure~\ref{fig:prompt-image}).

\subsection{Rolling Summary Prompts}
\label{prompt:rolling-summary}

Under \textbf{Rolling Summary}, the pre-decision EHR is split into consecutive 7-day windows.
For each non-final window, the model receives the previous cumulative summary and the current window's EHR events, and outputs an updated cumulative summary.
On the final window, the model receives the cumulative summary, the last window's EHR events, and the question, and outputs the MCQ answer in JSON format.

\subsection{Agent Prompt}
\label{app:prompts-agent}

Under the \textbf{History-level EHR} setting, \textbf{Agent} answers each MCQ through a multi-turn tool-calling loop.
The agent has three tools: \texttt{get\_overview} for patient overview, \texttt{search\_ehr} for semantic retrieval over the full pre-decision EHR, and \texttt{submit\_answer} for submitting the final option and rationale.
Only records on or before the decision cutoff are accessible.

\subsection{Evidence Utilization Judge Prompt}
\label{app:prompts-judge}

For the context-length analysis in Section~\ref{sec:context-length}, we use an LLM judge to score how substantively each model's rationale draws on the gold supporting evidence $\mathcal{E}$.
The judge receives the formatted question, the filtered numbered evidence list (using the same cutoff and scrub rules as \textbf{Evidence-only}), and the candidate model's \texttt{chosen\_option} and \texttt{reasoning}.
Unlike the MCQ prompts above, this judge prompt was deployed in English (Figure~\ref{fig:prompt-judge}).

\clearpage
\onecolumn
\captionsetup[figure]{skip=3pt}
\setlength{\intextsep}{5pt}

\begin{table}[H]
  \centering
  \footnotesize
  \setlength{\tabcolsep}{5pt}
  \renewcommand{\arraystretch}{1.08}
  \begin{tabularx}{\textwidth}{|>{\raggedright\arraybackslash\bfseries}p{3.5cm}|>{\raggedright\arraybackslash}X|>{\raggedright\arraybackslash}p{3.1cm}|}
    \hline
    \rowcolor{gray!17}
    \multicolumn{3}{|c|}{\textbf{Instantiations of the MCQ Inference Prompt}} \\
    \hline
    \rowcolor{gray!8}
    \textbf{Setting} & \textbf{Scope instruction} & \textbf{User input header} \\
    \hline
    Evidence-only &
    The following text contains \textbf{key clinical evidence} extracted from the patient's EHR. These items were curated by another physician reviewing the record and cover the information relevant to the current decision. Answer using only these evidence items and the question. &
    Key Clinical Evidence \\
    \hline
    Visit-level EHR &
    The following text contains the patient's basic profile and \textbf{EHR events from the most recent 1 day}. Earlier records are omitted. Answer using only the provided information and the question. &
    EHR Excerpt \\
    \hline
    History-level EHR &
    Answer using only the provided \textbf{EHR excerpt} and the question. &
    EHR Excerpt \\
    \hline
    Static RAG / Static RAG + Rec. &
    The following text contains the \textbf{record snippets retrieved} from the full EHR that are most relevant to the question. The full history may contain additional information, but only the retrieved snippets are shown. If the retrieved information is insufficient, make the best decision based on what is available. &
    See Fig.~\ref{fig:prompt-static-rag} \\
    \hline
    Image &
    The following \textbf{images} contain the patient's EHR records (text rendered as page-level images). Read all images carefully and answer the question based on the clinical content. &
    See Fig.~\ref{fig:prompt-image} \\
    \hline
  \end{tabularx}
  \caption{Instantiations of the MCQ inference prompt across input-scope and access settings. Static RAG + Recency uses the same prompt as Static RAG; only the retrieval weighting differs. Image uses a vision-language multi-modal user message.}
  \label{tab:prompt-variations}
\end{table}

\begin{figure}[H]
  \centering
  \begin{promptbox}{Decision Chain Discovery Prompt (English Translation)}
    \textbf{System message}\par\smallskip
    You are an expert in obstetric clinical guideline analysis. Your task is to systematically discover all \textbf{clinical decision chains} from textbook text---patterns of the form ``if a pregnant patient presents condition X $\rightarrow$ management action Y should be performed within time T.''

    Core principles:
    \begin{enumerate}\setlength{\itemsep}{2pt}\setlength{\parskip}{0pt}
      \item Extract only from the given text; do not add external knowledge.
      \item Each chain must contain an explicit trigger-condition $\rightarrow$ management-action causal relation.
      \item There must be a temporal span (trigger and action are separated by days to months, not instantaneous operations).
      \item Preserve numeric thresholds from the source text (e.g., ``fasting glucose $\geq$5.1\,mmol/L'') and medication doses.
      \item Identify all branch decision points---under the same trigger, different situations follow different management paths.
      \item Do not miss chains: one chapter may contain multiple distinct decision chains.
    \end{enumerate}

    Granularity control (very important):
    \begin{itemize}\setlength{\itemsep}{2pt}\setlength{\parskip}{0pt}
      \item Each chain corresponds to one independently answerable \textbf{core clinical question}, e.g., ``After GDM diagnosis, is full-pregnancy management needed?''
      \item Do not split sub-steps of the same management action into separate chains; merge them as branches under one chain.
      \item Independent decisions across different time windows should remain separate chains.
      \item Prenatal management, delivery-mode decisions, and postpartum follow-up are usually three independent chains.
    \end{itemize}

    \smallskip
    \textbf{User message}\par\smallskip
    Discover all clinical decision chains from the following textbook/guideline text.

    Chapter/guideline name: \textit{\{CHAPTER\_NAME\}}\\
    Page range: \textit{\{PAGE\_RANGE\}}

    \textit{\{HINT\_SECTION\}}

    \#\# Text content\\
    \textit{\{OCR\_TEXT\}}

    Output JSON listing all discovered chains with fields such as \texttt{chain\_id}, \texttt{trigger\_condition}, \texttt{trigger\_criteria}, \texttt{management\_action}, \texttt{expected\_timeline}, \texttt{branch\_points}, \texttt{confidence}, and \texttt{textbook\_quotes}.
    A chapter typically contains 3--8 chains; each chain should have at least two branch points, including a negative ``not found/not performed'' branch.
  \end{promptbox}
  \caption{Prompt for discovering clinical decision chains from textbook/guideline OCR text.}
  \label{fig:prompt-chain-discovery}
\end{figure}

\begin{figure}[H]
  \centering
  \begin{promptbox}{Candidate Branch Classification Prompt (English Translation)}
    \textbf{System message}\par\smallskip
    You are a senior obstetric clinical expert agent. Based on the patient's EMR data, determine which branch of the specified clinical rule applies.

    Workflow:
    \begin{enumerate}\setlength{\itemsep}{2pt}\setlength{\parskip}{0pt}
      \item Review basic patient information.
      \item Read key data targeted by the rule trigger and medical background.
      \item Request additional record slices if needed.
      \item Submit a branch classification when sufficient evidence is available.
    \end{enumerate}

    \#\# Clinical rule: \textit{\{CATEGORY\_NAME\}}

    \#\# Medical background\\
    \textit{\{MEDICAL\_BACKGROUND\}}

    \#\# Candidate branches (choose exactly one branch ID; do not invent new IDs)\\
    \textit{\{BRANCHES\_DESC\}}

    \#\# Valid branch ID list\\
    \textit{\{BRANCH\_IDS\_JSON\}}

    Important notes:
    \begin{itemize}\setlength{\itemsep}{2pt}\setlength{\parskip}{0pt}
      \item The \texttt{branch} field must be one of the valid branch IDs above; do not use Chinese labels or custom IDs.
      \item For hypertension-related rules, even a single abnormal BP reading together with a relevant diagnosis may support classification.
      \item Admission/discharge diagnoses are valid clinical evidence.
    \end{itemize}

    \smallskip
    \textbf{User message}\par\smallskip
    Patient summary and candidate metadata: \textit{\{PATIENT\_SUMMARY\}}, \textit{\{RULE\_ID\}}, \textit{\{PATIENT\_ID\}}.
  \end{promptbox}
  \caption{Core system prompt for agent-based candidate branch classification after programmatic screening.}
  \label{fig:prompt-candidate-classification}
\end{figure}

\begin{figure}[H]
  \centering
  \begin{promptbox}{Chain Audit Prompt (English Translation)}
    \textbf{System message}\par\smallskip
    You are an obstetric clinical item-review expert. Determine whether the case's reasoning chain is sufficient to support construction of a medical reasoning question.

    Criteria:
    \begin{enumerate}\setlength{\itemsep}{2pt}\setlength{\parskip}{0pt}
      \item Do \texttt{key\_events} form an interpretable medical reasoning chain?
      \item Even if retrospective information is mixed in, can the case still be rewritten as a valid current-time reasoning question?
      \item Is there an obvious scene mismatch between the rule/category and \texttt{key\_events} (e.g., preterm rule on a term case, syphilis rule without syphilis evidence, tocolysis rule used for antihypertensive management)?
    \end{enumerate}

    Output JSON: \texttt{\{"audit\_pass": bool, "audit\_reason": str, "focus\_terms": list\}}.

    Set \texttt{audit\_pass=false} only when:
    \begin{itemize}\setlength{\itemsep}{2pt}\setlength{\parskip}{0pt}
      \item the chain is medically invalid;
      \item the branch strongly conflicts with \texttt{key\_events};
      \item the rule/case scene is clearly mismatched;
      \item no valid current-time question can be inferred.
    \end{itemize}
    Do not generate a question; do not output extra fields.

    \smallskip
    \textbf{User message}\par\smallskip
    \textit{\{RULE\_ID, CATEGORY\_NAME, BRANCH, QUESTION\_TEMPLATE, MEDICAL\_BACKGROUND, EVIDENCE\_SUMMARY, KEY\_EVENTS, TOOL\_CALLS, TRIGGER\_EVENTS\_DATES\}}
  \end{promptbox}
  \caption{Prompt for auditing whether a candidate case supports question construction.}
  \label{fig:prompt-chain-audit}
\end{figure}

\begin{figure}[H]
  \centering
  \begin{promptbox}{Draft Question Generation Prompt (English Translation)}
    \textbf{System message}\par\smallskip
    You are an obstetric clinical item-writing expert. Draft a medical reasoning question based on the case information.

    Core principles:
    \begin{enumerate}\setlength{\itemsep}{2pt}\setlength{\parskip}{0pt}
      \item Test clinical reasoning and decision-making, not static label classification.
      \item The stem gives a clinical scenario and a decision question; the test-taker must look up evidence in the EHR.
      \item The semantic direction of \texttt{gold\_option} is locked by \texttt{branch\_label} and must not change.
    \end{enumerate}

    Stem constraints:
    \begin{itemize}\setlength{\itemsep}{2pt}\setlength{\parskip}{0pt}
      \item Must bind the exact \texttt{cutoff} date as the decision time point.
      \item Use neutral background wording (e.g., ``follow-up in a high-risk obstetric clinic''); do not translate disease names from the rule/category.
      \item Do not include specific lab values, specific diagnoses, exclusion statements (``no XX seen''), or full abnormal chains.
      \item Do not use meta-conclusion phrasing such as ``already recorded/already decided''.
      \item Do not translate EHR key findings line-by-line into the stem.
    \end{itemize}

    Option constraints:
    \begin{itemize}\setlength{\itemsep}{2pt}\setlength{\parskip}{0pt}
      \item Correct and distractor options must sit at the same abstraction level.
      \item Write options as clinical decisions, not retrieval-style wording such as ``evidence not found''.
      \item Distractors should resemble realistic alternative management choices.
    \end{itemize}

    Output JSON: \texttt{\{"draft\_question\_stem", "options" [4], "gold\_option", "grounding\_focus"\}}; if a valid item cannot be written, return \texttt{\{"reject\_reason"\}} with other fields empty.

    \smallskip
    \textbf{User message}\par\smallskip
    \textit{\{RULE\_ID, CATEGORY\_NAME, BRANCH, BRANCH\_LABEL, QUESTION\_TEMPLATE, BRANCHES, EVIDENCE\_SUMMARY, KEY\_EVENTS, TOOL\_CALLS, AUDIT\_REASON, FOCUS\_TERMS\}}
  \end{promptbox}
  \caption{Prompt for drafting an examination-style clinical question and four options.}
  \label{fig:prompt-draft-question}
\end{figure}

\begin{figure}[H]
  \centering
  \begin{promptbox}{Grounding Verification Prompt (English Translation)}
    \textbf{System message}\par\smallskip
    You are an obstetric clinical evidence-verification agent. Take the draft question into the EHR for targeted verification.

    \#\# Verification steps
    \begin{enumerate}\setlength{\itemsep}{2pt}\setlength{\parskip}{0pt}
      \item \textbf{First step (required):} verify the stem premise---confirm that the EHR contains original evidence for the core disease implied by the rule. If not found (e.g., all syphilis tests negative, bile acids never abnormal) $\rightarrow$ \texttt{grounded\_pass=false}. Upstream \texttt{key\_events} may be wrong; do not trust them blindly.
      \item Verify that the gold answer is supported by original record evidence.
      \item Select a \texttt{cutoff} date:
      \begin{itemize}\setlength{\itemsep}{1pt}\setlength{\parskip}{0pt}
        \item Using only data on or before \texttt{cutoff}, the test-taker must be able to derive the gold answer.
        \item Every \texttt{support\_evidence} item must have date $\leq$ \texttt{cutoff}.
        \item All post-\texttt{cutoff} data are hidden from the test-taker; set \texttt{cutoff} at the end of the evidence chain, not the beginning.
        \item If repeated measurements are needed, \texttt{cutoff} $\geq$ the date of the last key repeat measurement.
        \item For inpatient cases, set \texttt{cutoff} to the admission date (the program automatically removes all diagnoses on that day).
        \item If no \texttt{cutoff} can satisfy both answerability and non-leakage, \texttt{grounded\_pass=false}.
      \end{itemize}
      \item Generate \texttt{precise\_scrub\_targets} (exact deletion list).
      \item If verification cannot be completed within the tool-call budget, \texttt{grounded\_pass=false}.
    \end{enumerate}

    \#\# Scrub rules
    \begin{itemize}\setlength{\itemsep}{2pt}\setlength{\parskip}{0pt}
      \item After choosing \texttt{cutoff}, the program automatically removes all diagnoses on that day; notes/procedures on that day that leak the answer must be listed for scrubbing.
      \item Post-\texttt{cutoff} data are globally masked; do not list them in scrub targets.
      \item Do not delete pre-\texttt{cutoff} labs or vitals.
      \item Pre-\texttt{cutoff} notes/medications may be deleted only when they state management conclusions or discharge summaries.
      \item Labs on \texttt{cutoff} day are kept by default; same-day follow-up tests (e.g., OGTT 1h/2h) may be scrubbed with reason prefixed by ``[same-day follow-up]'' and no conditional wording.
      \item \texttt{match\_hint} contains only \texttt{value\_text}, without a name prefix.
      \item Scrub only answer leakage, retrospective outcomes, or management conclusions; do not output spurious scrub items.
    \end{itemize}

    \#\# Gold direction
    \begin{itemize}\setlength{\itemsep}{2pt}\setlength{\parskip}{0pt}
      \item \texttt{branch\_label} locks the semantic direction of the gold answer; it must not change.
      \item If the EHR does not support the branch, set \texttt{grounded\_pass=false} with \texttt{drop\_reason}; do not invent a new direction.
      \item You do not assign L1/L2/L3; the level is determined by the rule system.
    \end{itemize}

    \smallskip
    \textbf{User message}\par\smallskip
    Draft question, options, gold option, grounding focus, audit reason, focus terms, key events, and trigger-event dates: \textit{\{CASE\_AND\_DRAFT\_FIELDS\}}.
  \end{promptbox}
  \caption{Core system prompt for grounding verification: cutoff selection, supporting evidence collection, and leakage scrubbing.}
  \label{fig:prompt-grounding}
\end{figure}

\begin{figure}[H]
  \centering
  \begin{promptbox}{Final Question Rewrite Prompt (English Translation)}
    \textbf{System message}\par\smallskip
    You are an obstetric clinical item editor. Without changing the core testing point or answer, revise the draft into the final item.

    Core principles:
    \begin{enumerate}\setlength{\itemsep}{2pt}\setlength{\parskip}{0pt}
      \item Preserve medical-reasoning item style; the stem may use only information supported by \texttt{support\_evidence}.
      \item The semantic direction of \texttt{gold\_option} is locked by \texttt{branch\_label} and must not change.
      \item Do not introduce new facts or change option abstraction level.
    \end{enumerate}

    Stem constraints:
    \begin{itemize}\setlength{\itemsep}{2pt}\setlength{\parskip}{0pt}
      \item Explicitly state \texttt{confirmed\_cutoff\_date} as the decision time point.
      \item Use neutral background wording; do not translate disease names from rule labels.
      \item Do not include specific values, specific diagnoses, exclusion statements, full abnormal chains, or meta-conclusion phrasing.
      \item Keep the stem short and examination-like; the test-taker must consult the EHR to answer.
    \end{itemize}

    Option constraints:
    \begin{itemize}\setlength{\itemsep}{2pt}\setlength{\parskip}{0pt}
      \item Write options as clinical decisions, not retrieval-style wording.
      \item Do not change gold semantics; only improve natural phrasing.
    \end{itemize}

    Output JSON: \texttt{\{"final\_question\_stem", "final\_options" [4], "final\_gold\_option"\}}.

    \smallskip
    \textbf{User message}\par\smallskip
    \textit{\{CONFIRMED\_CUTOFF\_DATE, BRANCH\_LABEL, DRAFT\_STEM, DRAFT\_OPTIONS, DRAFT\_GOLD, SUPPORT\_EVIDENCE, REWRITE\_NOTES\}}
  \end{promptbox}
  \caption{Prompt for rewriting a grounded draft into the final benchmark question.}
  \label{fig:prompt-final-rewrite}
\end{figure}

\begin{figure}[H]
  \centering
  \begin{promptbox}{Shared MCQ Inference Prompt (English Translation)}
    \textbf{System message}\par\smallskip
    You are a senior obstetric attending physician. \textit{\{SCOPE\_INSTRUCTION\}}

    Your output must be a single-line JSON object without markdown code fences. Keys:
    \begin{itemize}\setlength{\itemsep}{2pt}\setlength{\parskip}{0pt}
      \item \texttt{chosen\_option}: string ``A'' $|$ ``B'' $|$ ``C'' $|$ ``D''
      \item \texttt{reasoning}: brief rationale in Chinese
    \end{itemize}

    \smallskip
    \textbf{User message}\par\smallskip
    \#\# \textit{\{USER\_INPUT\_HEADER\}}\\
    \textit{\{CLINICAL\_INPUT\}}

    \#\# Question\\
    \textit{\{FORMATTED\_QUESTION\}}

    Please output JSON.
  \end{promptbox}
  \caption{Shared MCQ inference prompt. The system message is reused by all settings in Table~\ref{tab:prompt-variations}; this figure shows the text-based user-message layout used by Evidence-only, Visit-level EHR, and History-level EHR.}
  \label{fig:prompt-mcq-shared}
\end{figure}

\begin{figure}[H]
  \centering
  \begin{promptbox}{Static RAG MCQ Prompt (English Translation)}
    \textbf{System message}\par\smallskip
    You are a senior obstetric attending physician. \textit{\{SCOPE\_INSTRUCTION\}}

    Your output must be a single-line JSON object without markdown code fences. Keys: \texttt{chosen\_option} (``A''$|$``B''$|$``C''$|$``D''), \texttt{reasoning} (brief rationale in Chinese).

    \smallskip
    \textbf{User message}\par\smallskip
    \#\# Patient Overview\\
    \textit{\{PATIENT\_OVERVIEW\}}

    \#\# Retrieved Record Snippets (\textit{\{K\}} snippets, ranked by relevance)\\
    1. \textit{\{SNIPPET\_1\}}\\
    2. \textit{\{SNIPPET\_2\}}\\
    \textit{...}

    \#\# Question\\
    \textit{\{FORMATTED\_QUESTION\}}

    Please output JSON.
  \end{promptbox}
  \caption{MCQ prompt for Static RAG and Static RAG + Recency. The scope instruction is instantiated in Table~\ref{tab:prompt-variations}.}
  \label{fig:prompt-static-rag}
\end{figure}

\begin{figure}[H]
  \centering
  \begin{promptbox}{Image MCQ Prompt (English Translation)}
    \textbf{System message}\par\smallskip
    You are a senior obstetric attending physician. \textit{\{SCOPE\_INSTRUCTION\}}

    Your output must be a single-line JSON object without markdown code fences. Keys: \texttt{chosen\_option} (``A''$|$``B''$|$``C''$|$``D''), \texttt{reasoning} (brief rationale in Chinese).

    \smallskip
    \textbf{User message (multi-modal)}\par\smallskip
    \textit{\{RENDERED\_EHR\_PAGE\_1\}}\\
    \textit{\{RENDERED\_EHR\_PAGE\_2\}}\\
    \textit{...}

    \#\# Question\\
    \textit{\{FORMATTED\_QUESTION\}}

    Please output JSON.
  \end{promptbox}
  \caption{MCQ prompt for the Image access strategy. The scope instruction is instantiated in Table~\ref{tab:prompt-variations}. Each \textit{\{RENDERED\_EHR\_PAGE\_i\}} denotes one paginated PNG image of the pre-decision EHR text.}
  \label{fig:prompt-image}
\end{figure}

\begin{figure}[H]
  \centering
  \begin{promptbox}{Rolling Summarization Prompt (English Translation)}
    \textbf{System message}\par\smallskip
    You are a senior obstetric attending physician, reading a patient's EHR in segments. Based on the \textbf{previous summary} and the \textbf{new records in the current segment}, output an updated \textbf{cumulative summary} that includes:
    \begin{enumerate}\setlength{\itemsep}{2pt}\setlength{\parskip}{0pt}
      \item Key personal information (age, gravidity/parity, BMI, important history, etc.)
      \item Important clinical indicators and trends (e.g., blood pressure, glucose, weight, laboratory results)
      \item Abnormal indicators and confirmed or suspected diagnoses
      \item Current medications and important procedures
    \end{enumerate}
    The summary should be concise yet complete, and should not omit important information recorded in earlier segments.

    \smallskip
    \textbf{User message}\par\smallskip
    \#\# Patient Overview\\
    \textit{\{PATIENT\_OVERVIEW\}}

    \#\# Previous Cumulative Summary\\
    \textit{\{PREVIOUS\_SUMMARY\}} \hfill (omitted in the first window)

    \#\# Segment \textit{\{i\}} Records (\textit{\{DATE\_RANGE\}}, \textit{\{N\}} events)\\
    \textit{\{CURRENT\_WINDOW\_EHR\}}

    Please output the updated cumulative summary.
  \end{promptbox}
  \caption{Rolling summarization prompt used for all non-final 7-day windows.}
  \label{fig:prompt-rolling-summarize}
\end{figure}

\begin{figure}[H]
  \centering
  \begin{promptbox}{Rolling Summary Final MCQ Prompt (English Translation)}
    \textbf{System message}\par\smallskip
    You are a senior obstetric attending physician. You have read the patient's full EHR in segments. The following text contains the cumulative summary from earlier segments and the records from the final segment. Answer the question based on all provided information.

    Your output must be a single-line JSON object without markdown code fences. Keys: \texttt{chosen\_option} (``A''$|$``B''$|$``C''$|$``D''), \texttt{reasoning} (brief rationale in Chinese).

    \smallskip
    \textbf{User message}\par\smallskip
    \#\# Previous Cumulative Summary\\
    \textit{\{CUMULATIVE\_SUMMARY\}}

    \#\# Final Segment Records (\textit{\{DATE\_RANGE\}}, \textit{\{N\}} events)\\
    \textit{\{FINAL\_WINDOW\_EHR\}}

    \#\# Question\\
    \textit{\{FORMATTED\_QUESTION\}}

    Please output JSON.
  \end{promptbox}
  \caption{Final-window MCQ prompt used in the Rolling Summary access strategy.}
  \label{fig:prompt-rolling-final}
\end{figure}

\begin{figure}[H]
  \centering
  \begin{promptbox}{Agent Prompt (English Translation)}
    \textbf{System message}\par\smallskip
    You are a clinical physician answering a multiple-choice question based on the patient's EHR. You may use the following tools to consult the record:
    \begin{itemize}\setlength{\itemsep}{2pt}\setlength{\parskip}{0pt}
      \item \texttt{get\_overview}: retrieve the patient overview
      \item \texttt{search\_ehr}: search EHR snippets by keyword (may be called multiple times with different queries)
      \item \texttt{submit\_answer}: submit the final answer
    \end{itemize}

    Note: only records on or before \textit{\{CUTOFF\_DATETIME\}} (inclusive) may be used.
    First review the patient overview, then search for information needed by the question, and finally call \texttt{submit\_answer}.

    \smallskip
    \textbf{Initial user message}\par\smallskip
    \textit{\{FORMATTED\_QUESTION\}}

    Please first call \texttt{get\_overview} to review the patient's basic information.

    \smallskip
    \textbf{Tool: \texttt{submit\_answer}}\par\smallskip
    Submit the final multiple-choice answer and rationale. Required fields: \texttt{chosen\_option} (``A''$|$``B''$|$``C''$|$``D''), \texttt{reasoning} (brief rationale in Chinese, citing key evidence).
  \end{promptbox}
  \caption{Initial Agent prompt used in the History-level EHR access strategy. Subsequent turns append tool responses and assistant tool calls until \texttt{submit\_answer} is invoked.}
  \label{fig:prompt-agent}
\end{figure}

\begin{figure}[H]
  \centering
  \begin{promptbox}{Evidence Utilization Judge Prompt}
    \textbf{System message}\par\smallskip
    You are an independent clinical case-item LLM judge.

    Your task: using ONLY the question stem/options, the numbered key clinical evidence list, and the candidate's reasoning text below, rate how substantively the candidate's \textbf{reasoning draws on items in that evidence list} (explicitly or implicitly---labs, course, orders, dates, etc.).

    Rules:
    \begin{itemize}\setlength{\itemsep}{2pt}\setlength{\parskip}{0pt}
      \item Do NOT require the candidate to restate every evidence item; focus on decision-relevant evidence in the reasoning chain.
      \item If the filtered evidence count is 0, assign score \textbf{0} per the user message.
      \item Output a single JSON object (no markdown fences) parseable by \texttt{json.loads}.
      \item \texttt{support\_evidence\_utilization\_score} must be an integer \textbf{0--5}.
      \item All string fields in the JSON must be in \textbf{English}.
    \end{itemize}

    \smallskip
    \textbf{User message}\par\smallskip
    \#\# Question stem and options\\
    \textit{\{FORMATTED\_QUESTION\}}

    \#\# Key clinical evidence for this item (post-filter; same as at exam time; \textbf{\textit{\{N\_EVIDENCE\}}} items)\\
    Indices match the numbered list below. In \texttt{matched\_evidence\_indices}, list \textbf{1-based} indices of items substantively used in the reasoning.

    \textit{\{FILTERED\_SUPPORT\_EVIDENCE\}}

    ---

    \#\# Candidate response (model under evaluation)\\
    chosen\_option: \textit{\{CHOSEN\_OPTION\}}\\
    reasoning:\\
    \textit{\{CANDIDATE\_REASONING\}}

    ---

    \#\# Rubric: support\_evidence\_utilization\_score (integer 0--5)
    \begin{itemize}\setlength{\itemsep}{2pt}\setlength{\parskip}{0pt}
      \item \textbf{5}: Reasoning clearly weaves \textbf{multiple decision-relevant} evidence items into the main argument; specific clinical facts tie to the conclusion (not vague platitudes).
      \item \textbf{4}: \textbf{Most} relevant items are used; strong alignment with the list; minor items may be omitted.
      \item \textbf{3}: \textbf{Some} items used, but important ones missing or only generic references (``labs'', ``chart'') without substantive alignment.
      \item \textbf{2}: Touches only one minor point, or relies on inference poorly aligned with the list.
      \item \textbf{1}: Almost no substantive use of the list (or contradicts evidence while forcing a choice).
      \item \textbf{0}: No use of the list (generic/off-chart reasoning), or \textit{\{N\_EVIDENCE\}} = 0 and reasoning does not reasonably ground on the stem / is pure speculation.
    \end{itemize}

    If \textit{\{N\_EVIDENCE\}} = 0: \texttt{support\_evidence\_utilization\_score} must be \textbf{0}; \texttt{matched\_evidence\_indices} = \texttt{[]}; note in \texttt{rubric\_note} that no filtered evidence items were available.

    Return JSON with \textbf{only} these keys:
    \begin{itemize}\setlength{\itemsep}{2pt}\setlength{\parskip}{0pt}
      \item \texttt{support\_evidence\_utilization\_score}: integer 0--5
      \item \texttt{matched\_evidence\_indices}: array of integers (1-based indices used; \texttt{[]} if none)
      \item \texttt{rubric\_note}: string, one-sentence rationale ($\leq$200 chars, English)
      \item \texttt{summary}: string, 2--5 sentences in English on alignment between reasoning and evidence
    \end{itemize}
  \end{promptbox}
  \caption{LLM-as-judge prompt for scoring evidence utilization in model rationales. \textit{\{FILTERED\_SUPPORT\_EVIDENCE\}} is built with the same cutoff and scrub filtering as the \textbf{Evidence-only} input.}
  \label{fig:prompt-judge}
\end{figure}

\end{document}